\documentclass{article}

\usepackage[preprint,nonatbib]{neurips_2020}

\usepackage[utf8]{inputenc} 
\usepackage[T1]{fontenc}    
\usepackage{url}            
\usepackage{booktabs}       
\usepackage{amsfonts}       
\usepackage{nicefrac}       
\usepackage{microtype}      
\usepackage{graphicx}
\usepackage{mathtools}
\usepackage{amsmath}
\usepackage{amssymb}
\usepackage{wrapfig}
\usepackage[mathscr]{eucal}
\usepackage[pagebackref=true,colorlinks,bookmarks=false,citecolor=blue,linkcolor=blue]{hyperref}
\usepackage[dvipsnames,svgnames,x11names]{xcolor}

\newcommand{\norm}[1]{\left\lVert#1\right\rVert}

\newcommand{\specialcell}[2][c]{%
  \begin{tabular}[#1]{@{}c@{}}#2\end{tabular}}

\title{Online Adaptation for Consistent Mesh Reconstruction in the Wild}

\author{
Xueting Li$^{1}$, 
 \textbf{Sifei Liu$^{2}$}, \
 \textbf{Shalini De Mello$^{2}$}, \
 \textbf{Kihwan Kim$^{2}$}, \\
 \textbf{Xiaolong Wang$^{3}$}, \
 \textbf{Ming-Hsuan Yang}$^{1}$, \
 \textbf{Jan Kautz}$^{2}$
\\ $^{1}$University of California, Merced, \
 $^{2}$NVIDIA, \
 $^{3}$University of California, San Diego}

\begin{document}
\maketitle

\begin{abstract}
This paper presents an algorithm to reconstruct temporally consistent 3D meshes of deformable object instances from videos in the wild.
Without requiring annotations of 3D mesh, 2D keypoints, or camera pose for each video frame, we pose video-based reconstruction as a self-supervised online adaptation problem applied to any incoming test video.
We first learn a category-specific 3D reconstruction model from a collection of single-view images of the same category that jointly predicts the shape, texture, and camera pose of an image.
Then, at inference time, we adapt the model to a test video over time using self-supervised regularization terms that exploit  temporal consistency of an object instance to enforce that all reconstructed meshes share a common texture map, a base shape, as well as parts.
We demonstrate that our algorithm recovers temporally consistent and reliable 3D structures from videos of non-rigid objects including those of animals captured in the wild -- an extremely challenging task rarely addressed before. 
Codes and other resources will be maintained at \url{https://sites.google.com/nvidia.com/vmr-2020}.
\end{abstract}

\section{Introduction}
\label{sec:intro}

\begin{figure}
  \centering
    \includegraphics[width=0.9\textwidth]{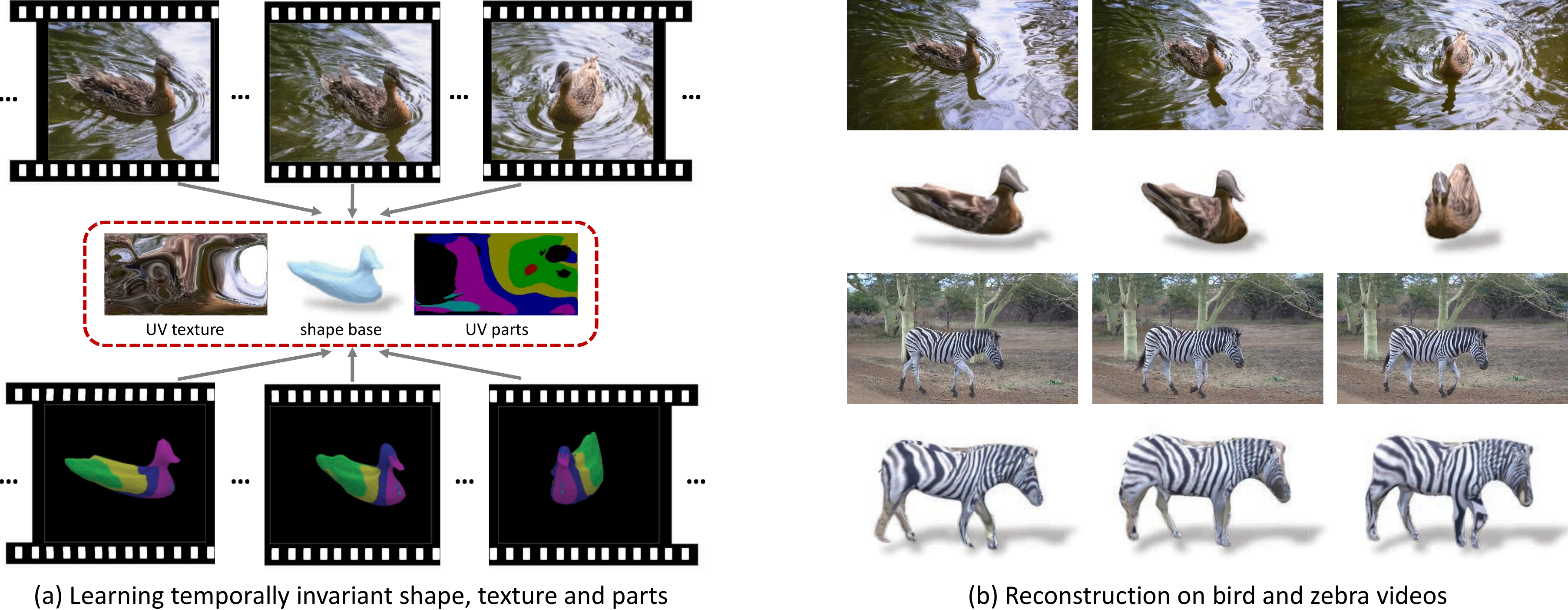}
  \caption{\footnotesize By utilizing the consistency of texture, shape and object parts correspondences in videos (red box) as self-supervision signals in (a), we learn a model that reconstructs temporally consistent meshes of deformable object instances in videos in (b).}\label{fig:teaser}
  \vspace{-6pt}
\end{figure}

When we humans try to understand the object shown in Fig.~\ref{fig:teaser}(a), we instantly recognize it as a ``duck''. We also instantly perceive and imagine its shape in the 3D world, its viewpoint, and its appearance from other views. Furthermore, when we see it in a video, its 3D structure and deformation become even more apparent to us. Our ability to perceive the 3D structure of objects contributes vitally to our rich understanding of them.

While 3D perception is easy for humans, 3D reconstruction of deformable objects remains a very challenging problem in computer vision, especially for objects in the wild. For learning-based algorithms, the key bottleneck is the lack of supervision. It is extremely challenging to collect 3D annotations such as 3D shape and camera pose~\cite{choy20163d,kato2018renderer}. Consequently, existing research mostly focuses on limited domains (e.g., rigid objects~\cite{lin2019photometric}, human bodies~\cite{kanazawa2019learning,zhang2019predicting} and faces~\cite{Wu_2020_CVPR}) for which 3D annotations can be captured in constrained environments. However, these approaches do not generalize well to non-rigid objects captured in naturalistic environments (e.g., animals).
In non-rigid structure from motion methods~\cite{Bregler2000RecoveringN3,novotny2019c3dpo}, the 3D structure can be partially recovered from correspondences between multiple viewpoints, which are also hard to label. Due to constrained environments and limited annotations, it is nearly impossible to generalize these approaches to the 3D reconstruction of non-rigid objects (e.g., animals)  from images and videos captured in the wild.

Instead of relying on 3D supervision, weakly supervised or self-supervised approaches have been proposed for 3D mesh reconstruction. They use annotated 2D object keypoints~\cite{cmrKanazawa18}, category-level templates~\cite{kulkarni2019csm,kulkarni2020articulation} or silhouettes~\cite{umr2020}. However, to scale up learning with 2D annotations to hundreds of thousands of images is still non-trivial. 
This limits the generalization ability of current models to new domains. For example, a 3D reconstruction model trained on single-view images, e.g.,~\cite{cmrKanazawa18}, produces unstable and erratic predictions for video data. This is unsurprising, due to perturbations over time. However, the temporal signal in videos should provide us an advantage instead of a disadvantage, as recently shown on the task of optimizing a 3D \emph{rigid} object mesh w.r.t.\ a particular video~\cite{zhu2017object,lin2019photometric}. The question is, can we also take advantage of the redundancy in temporal sequences as a form of self-supervision in order to improve the reconstruction of dynamic 
non-rigid objects?


In this work, we address this problem with two important innovations.
First, we strike a balance between model generalization and specialization. That is, we train an image-based network on a set of images, while at test time we adapt it online to an input video of a particular instance.
Test-time training~\cite{sun2019test} is non-trivial since no labels are provided for the video.
The key is to introduce self-supervised objectives that can continuously improve the model.
To do so, we exploit the UV texture space, which provides a parameterization that is invariant to object deformation.
%
We encourage the sampled texture, as well as a group of object parts, to be consistent among all the individual frames in the UV space, as shown in Fig.~\ref{fig:teaser}(a).
Using this constraint of temporal consistency, the recovered shape and camera pose are stabilized considerably and are adapted to the current video.

One bottleneck of existing image-based 3D mesh reconstruction methods~\cite{cmrKanazawa18,umr2020} is that the predicted shapes are assumed to be symmetric.
This assumption does not hold for most non-rigid animals, e.g., birds tilting their heads, or walking horses, etc.
Our second innovation is to remove this assumption and to allow the reconstructed meshes to fit more complex, non-rigid poses via an as-rigid-as-possible (ARAP) constraint. 
As another constraint that does not require any labels, we enforce ARAP during test-time training as well, to substantially improve shape prediction.
We use two image-based 3D reconstruction models for training (i) a weakly supervised one (i.e., with object silhouettes and 2D keypoints provided), and (ii) a self-supervised one where only object silhouettes are available. The image-based models are then adapted to in-the-wild bird and zebra videos collected from the internet. 
We show that for both models, our innovations lead to an effective and robust approach to deformable, dynamic 3D object reconstruction of non-rigid objects captured in the wild.

\vspace{-0.10in}
\section{Related Work}
\vspace{-0.10in}
\label{sec:related}
\paragraph{3D object reconstruction from images.}
A triangular mesh has long been used for object reconstruction~\cite{cmrKanazawa18,kato2018renderer,liu2019softras,kato2019vpl,wang2018pixel2mesh,pan2019deep,wen2019pixel2mesh++}. It is a memory-efficient representation with vertices and faces, and is amenable to differentiable rendering techniques~\cite{kato2018renderer,liu2019softras}.
The task of 3D reconstruction entails the simultaneous recovery of the 3D shape, texture, and camera pose of objects from 2D images. It is highly ill-posed due to the inherent ambiguity of correctly estimating both the shape and camera pose together.
A major trend of recent works is to gradually reduce supervision from 3D vertices~\cite{choy20163d,wen2019pixel2mesh++,wang2018pixel2mesh}, shading~\cite{henderson2018learning}, or multi-view images~\cite{NIPS2016_6206,kato2018renderer,wiles2017silnet,rezende2016unsupervised,lin2019photometric} and move towards weakly supervised methods that instead use 2D semantic keypoints~\cite{cmrKanazawa18}, or a category-level 3D template~\cite{kulkarni2019csm}.
This progress makes the reconstruction of objects, e.g., birds, captured in the wild possible.
More recently, self-supervised methods~\cite{umr2020,Wu_2020_CVPR,kato2019self} have been developed to further remove the need for annotations.
Our method exploits different levels of
supervision: weak supervision (i.e., using 2D semantic keypoints) and self-supervision to learn an image-based 3D reconstruction network from a collection of images of a category (Sec.~\ref{sec:image-model}).

\paragraph{Non-rigid structure from motion (NR-SFM).}
NR-SFM aims to recover the pose and 3D structure of a non-rigid object, or object deforming non-rigidly over time, solely from 2D landmarks without 3D supervision~\cite{Bregler2000RecoveringN3}.
It is a highly ill-posed problem and needs to be regularized by additional shape priors~\cite{Bregler2000RecoveringN3,nrsfm14_zhu}.
Recently, deep networks~\cite{kong2019deep,novotny2019c3dpo} have been developed that serve as more powerful priors than the traditional approaches.
However, obtaining reliable landmarks or correspondences for videos is still a bottleneck. 
%
Our method bears resemblances to deep NR-SFM~\cite{novotny2019c3dpo}, which jointly predicts camera pose and shape deformation. Differently from them, we reconstruct dense meshes instead of sparse keypoints, without requiring labeled correspondences from videos.

\paragraph{3D object reconstruction from videos.}
%
Existing video-based object reconstruction methods mostly focus on specific domains, e.g., videos of faces~\cite{feng2018joint,tran2019learning} or human bodies~\cite{tung2017self,Arnab_CVPR_2019,doersch2019sim2real,kanazawa2019learning,zhang2019predicting}, where dense labelling is possible~\cite{Marcard_2018_ECCV}.
To augment video labels,~\cite{kanazawa2019learning} formulates dynamic human mesh reconstruction as an omni-supervision task, where a combination of labeled images and videos with pseudo-ground truth are used for training.
For human video-based 3D pose estimation, \cite{pavllo:videopose3d:2019} introduces semi-supervised learning to leverage unlabeled videos with a self-supervised component.
Dealing with specific application domains, all the aforementioned works rely on predefined shape priors, such as a parametric body model (e.g., SMPL~\cite{SMPL:2015}) or a morphable face model.
While our work also exploits unlabeled videos, we do not assume any predefined shape prior, which, practically, is hard to obtain for the majority of objects captured in the wild.

\paragraph{Optimization-based methods.}
Optimization-based methods have also been extensively explored for scene or object reconstruction from videos.
Several works~\cite{zhou2016sparseness,wandt20163d,rhodin2016general,2018-TOG-SFV} first obtain a single-view 3D reconstruction and then optimize the mesh and skeletal parameters.
Another line of methods is developed to optimize the weights of deep models instead, to render more robust results for a video of a particular instance ~\cite{tung2017self,lin2019photometric,Luo-VideoDepth-2020,Zuffi19Safari}.
Our method falls into this category.
While \cite{tung2017self} enforces consistency between observed 2D and a re-projection from 3D, \cite{lin2019photometric,Luo-VideoDepth-2020} take a further step and encourage consistency between frames via a network that inherently encodes an entire video into an invariant representation.
In this work, instead of limiting to rigid objects as~\cite{lin2019photometric}, or depth estimation as~\cite{Luo-VideoDepth-2020}, we recover dynamic meshes from videos captured in the wild -- a much more challenging problem that is rarely explored.  

\vspace{-0.10in}
\section{Approach}
\vspace{-0.10in}
\label{sec:method}

Our goal is to recover coherent sequences of mesh shapes, texture maps and camera poses from unlabeled videos, with a two-stage learning approach: (i) first, we learn a 3D mesh reconstruction model on a collection of single-view images of a category, described in Sec.~\ref{sec:image_model}; (ii) at inference time, we adapt the model to fit the sequence via temporal consistency constraints, as described in Sec.~\ref{sec:video_model}.
We focus on the weakly-supervised setting in Sec.~\ref{sec:image_model} and~\ref{sec:video_model}, where both silhouettes and keypoints are annotated in the image dataset. 
We then describe how to generalize the approach to a self-supervised setting, where only silhouettes are available in the image dataset in Sec.~\ref{sec:unsupervised}.
%
\paragraph{Notations.} We represent a textured mesh with $|V|$ vertices ($V\in{\mathbb{R}^{|V|\times 3}}$), $|F|$ faces ($F\in{\mathbb{R}^{|F|\times 3}}$) and a UV texture image ($I_\mathrm{uv}\in{\mathbb{R}^{H_\mathrm{uv}\times{W_\mathrm{uv}}\times 3}}$) of height $H_\mathrm{uv}$ and width $W_\mathrm{uv}$. 
Similarly to~\cite{cmrKanazawa18}, we use a weak perspective transformation to represent the camera pose $\theta\in\mathbb{R}^{7}$ of an input image. 
%
%
%
We denote $\mathcal{R}(\cdot)$ as a general projection, which can represent (i) a differentiable renderer~\cite{liu2019softras,kato2018renderer} to render a mesh to a 2D silhouette as $\mathcal{R}(V, \theta)$, or a textured mesh to an RGB image as $\mathcal{R}(V, \theta, I_\mathrm{uv})$ (we omit mesh faces $F$ for conciseness); (ii) or a projection of a 3D point $v$ to the image space as $\mathcal{R}(v, \theta)$. 
%
The Soft Rasterizer~\cite{liu2019softras} is used as the differentiable renderer in this work.

\vspace{-2mm}
\subsection{Single-view Mesh Reconstruction}
\vspace{-1mm}
\label{sec:image-model}
\label{sec:image_model}
\begin{figure}
  \centering
    \includegraphics[width=0.98\textwidth]{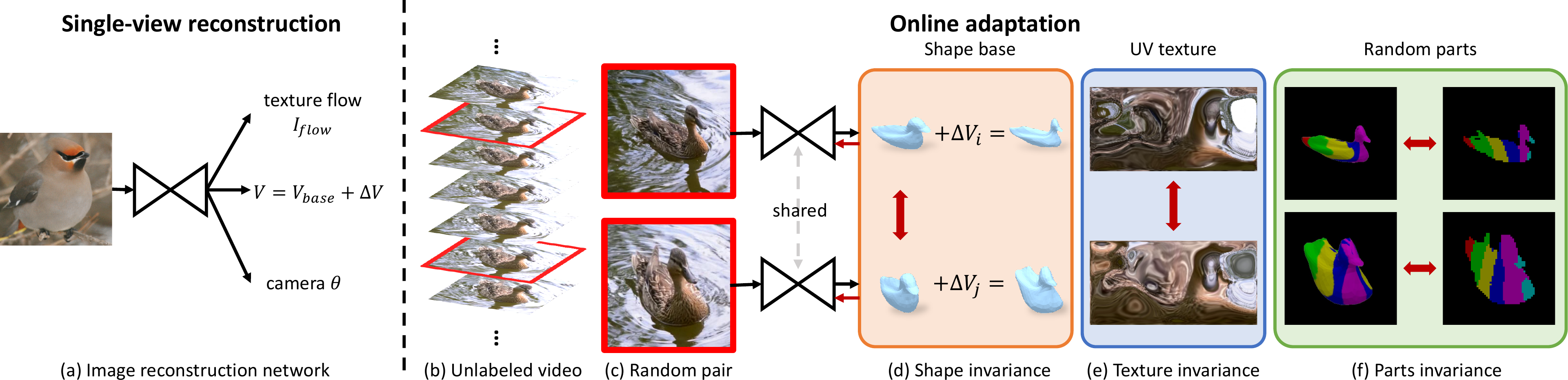}
  \caption{\footnotesize Overview. We show the single-view image reconstruction network on the left and the test-time training procedure to adapt it to a video on the right. Bold red arrows indicate invariance constraints in Sec.~\ref{sec:video_model}.}\label{fig:framework}
  \vspace{-7pt}
\end{figure}
In the first stage, 
we train a network with a collection of category-specific images that jointly estimates the shape, texture, and camera pose of an input image.
Similarly to~\cite{cmrKanazawa18}, we predict a texture flow $I_\mathrm{flow}\in{\mathbb{R}^{H_\mathbf{uv}\times{W_\mathbf{uv}}\times 2}}$ that maps pixels from the input image to the UV space. A predefined UV mapping function $\Phi$~\cite{hughes2014computer,cmrKanazawa18} is then used to map these pixels from the UV space to the mesh surface. 
%
%
With a differentiable renderer~\cite{liu2019softras}, we train the network with supervision from object silhouettes, texture, and the Laplacian objectives as in~\cite{cmrKanazawa18, umr2020}.
More details of learning texture and camera pose can be found in~\cite{cmrKanazawa18}.
An overview of our reconstruction framework is shown in Fig.~\ref{fig:framework}(a).

\paragraph{Recovering asymmetric shapes.} 
We propose a novel shape reconstruction module as shown in Fig.~\ref{fig:framework}(a). The key idea is to remove the symmetry requirement of object shapes, which is employed by many prior works~\cite{cmrKanazawa18,umr2020}. This is particularly important for recovering dynamic meshes in sequences, e.g., when a bird rotates its head as shown in Fig.~\ref{fig:img_quali}, its mesh is no longer mirror-symmetric.
Prior works~\cite{cmrKanazawa18,umr2020} model object shape by predicting vertex offsets from a jointly learned 3D template.
Simply removing the symmetry assumption for the predicted vertex offsets leads to excessive freedom in shape deformation, e.g., see Fig.~\ref{fig:img_quali}(f).
%
To resolve this, we learn a group of $N_b$ shape bases $\{{V_i}\}_{i=1}^{N_b}$, and replace the template by a weighted combination of them, denoted as the base shape $V_\mathrm{base}$.
Compared to a single mesh template, the base shape $V_\mathrm{base}$ is more powerful in capturing the object's identity and saves the model from predicting large motion deformation, e.g., of deforming a standing bird template to a flying bird.
The full shape reconstruction can be obtained by:
\vspace{-2mm}
\begin{align}
\label{eq:shape_base}
    V = V_\mathrm{base} + \Delta{V}, \quad \hspace{2mm} V_\mathrm{base} = \sum_{i=1}^{N_b}{{\beta}_{i}{V_i}},
    \vspace{-6mm}
\end{align}
where the $\Delta{V}$ encodes the object's asymmetric non-rigid motion and $\{{\beta}_{i}\}_{i=1}^{N_b}$ are learned coefficients.


%
The computation of our shape bases is inspired by parametric models~\cite{SMPL:2015,Zuffi:ICCV:2019,Zuffi:CVPR:2017}, where the basis shapes are extracted from an existing mesh dataset~\cite{CAESAR} or toy scans~\cite{Zuffi:CVPR:2017}. 
However, we make our model completely free of 3D supervision and obtain the bases by applying K-Means clustering to all meshes reconstructed by CMR~\cite{cmrKanazawa18}.
We use each cluster center as a basis shape in our model.
\paragraph{Keypoint re-projection.}

\begin{wrapfigure}{r}{0.6\textwidth}
   \vspace{-10pt}
  \centering
    \includegraphics[width=0.58\textwidth]{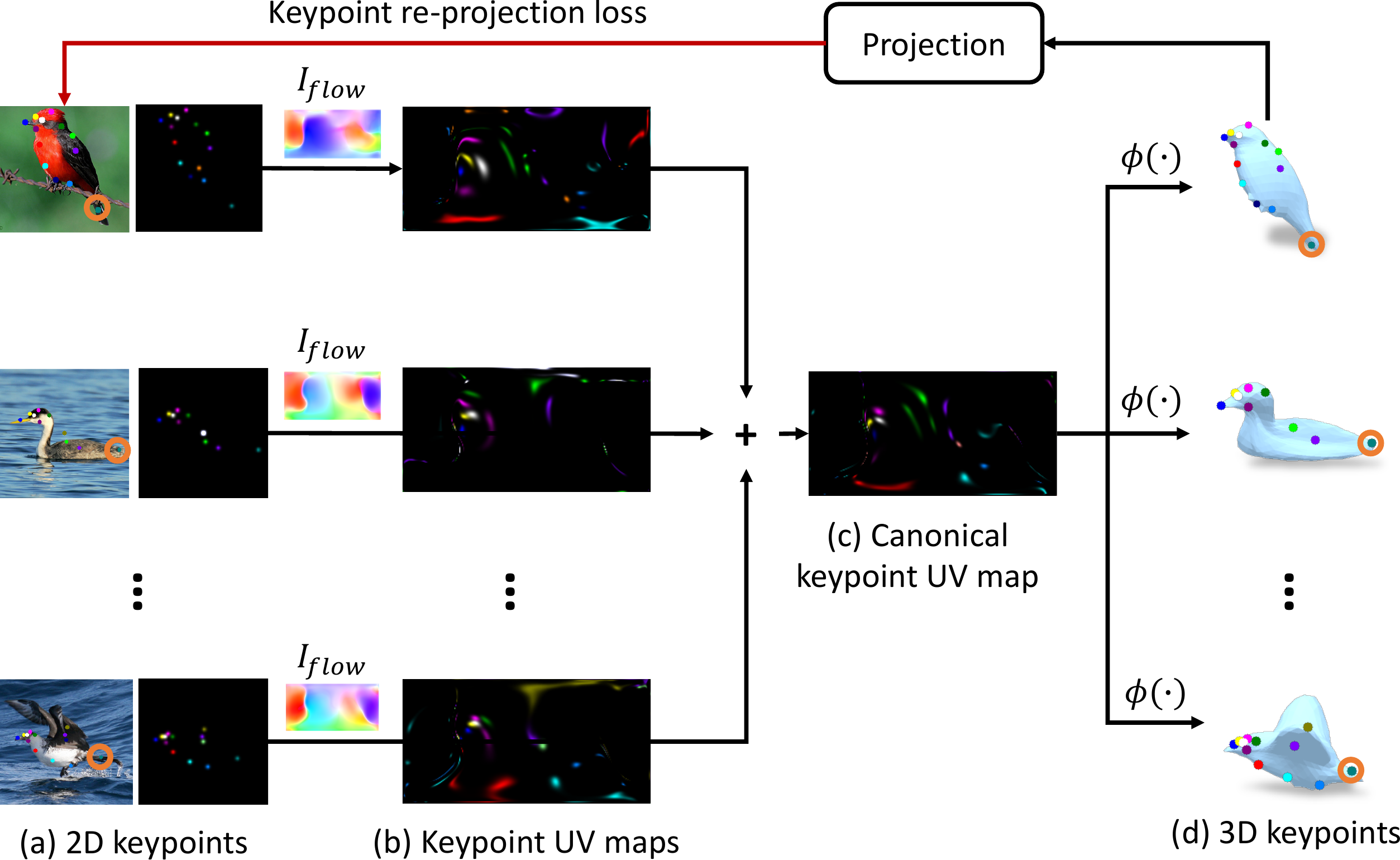}
  \caption{\footnotesize 3D canonical keypoints computation: (a) annotated 2D keypoints and their location heatmaps; (b) keypoint heatmaps mapped to the UV space using learned texture flows; (c) aggregated canonical keypoint heatmaps in the UV space; (d) canonical keypoints on different instance mesh surface. $\Phi(\cdot)$ is the UV mapping function discussed in Sec.~\ref{sec:image_model}.}\label{fig:kp_aggr}
  \vspace{-7pt}
\end{wrapfigure}

%
%
%

In the weakly-supervised setting, the 2D keypoints are provided that semantically associate different instances. When projected onto the mesh surface, the same semantic keypoint (e.g., the tail keypoint in the orange circle in Fig.~\ref{fig:kp_aggr}(a)) for different object instances should be matched to the same face on the mesh (the tail keypoint in the orange circle in Fig.~\ref{fig:kp_aggr}(d)). 
To model the mapping between the 3D mesh surface and the 2D keypoints, prior work~\cite{cmrKanazawa18} learns an affinity matrix that describes the probability of each 2D keypoint mapping to each vertex on the mesh.
The affinity matrix is shared among all instances and is independent of individual shape variations.
However, this approach is sub-optimal because:
(i) Mesh vertices are a subset of discrete points on a continuous mesh surface and so their weighted combination defined by the affinity matrix may not lie on it, leading to inaccurate mappings of 2D keypoints.
(ii) The mapping from the image space to the mesh surface described by the affinity matrix, in our case, however, is already modeled by the texture flow. Hence, it is potentially redundant to learn both of them independently.

In this work, we re-utilize texture flow to map 2D keypoints from each image to the mesh surface. 
%
We first map each 2D keypoint to the UV space that is independent of shape deformation (Fig.~\ref{fig:kp_aggr}(b)).
Ideally, each semantic keypoint from different instances should map to the same point in the UV space as discussed above. 
In practice, this does not hold due to inaccurate texture flow prediction.
To accurately map each keypoint to the UV space, we compute a canonical keypoint UV map as shown in Fig.~\ref{fig:kp_aggr}(c) by: (i) mapping the keypoint heat map in Fig.~\ref{fig:kp_aggr}(a) for each instance to the UV space via its predicted texture flow, and (ii) aggregating these keypoint UV maps in Fig.~\ref{fig:kp_aggr}(b) across all instances to eliminate outliers caused by incorrect texture flow prediction. 

We further utilize the pre-defined UV mapping function $\Phi$ discussed above to map each semantic keypoint from the UV space to the mesh surface.
Given the 3D correspondence (denoted as $K_{3D}^i$) of each 2D semantic keypoint $K_{2D}^i$, the keypoint re-projection loss enforces the projection of the former to be consistent with the latter by:
\vspace{-2mm}
\begin{equation}
L_{kp} = \frac{1}{N_k}\sum_{i=1}^{N_k}{\norm{\mathcal{R}(K_{3D}^i, \theta) - K_{2D}^i}},
\label{eq:kp_reproj}
\end{equation}
where $N_k$ is the number of keypoints.
%
%
\paragraph{As-rigid-as-possible (ARAP) constraint.}

%
Without any pose-related regularization, the predicted motion deformation $\Delta V$ often leads to erroneous random deformations and spikes as shown in Fig.~\ref{fig:img_quali}(f), which do not faithfully describe the motion of a non-rigid object.
Therefore, we introduce an as-rigid-as-possible (ARAP) constraint~\cite{sorkine2007rigid,deepcap} to encourage rigidity of local transformations and the preservation of the local mesh structure. 
Instead of solving the optimization in~\cite{sorkine2007rigid,deepcap}, we reformulate it as an objective that ensures that the predicted shape $V$ is a locally rigid transformation from the predicted base shape $V_\mathrm{base}$ by:
%
%
\vspace{-1mm}
\begin{equation}
L_\mathrm{arap}(V_\mathrm{base}, V) = \sum_{i=1}^{|V|}{\sum_{j\in{\mathcal{N}(i)}}{w_{ij}\norm{(V^i - V^j) - R_i({V^i_\mathrm{base}} - {V^j_\mathrm{base}})}}},
\label{eq:arap}
\end{equation}
where $\mathcal{N}(i)$ represents the neighboring vertices of a vertex $i$, $w_{ij}$ and $R_i$ are the cotangent weight and the best approximating rotation matrix, respectively, as described in~\cite{sorkine2007rigid}.


\vspace{-2mm}
\subsection{Online Adaptation for Videos}
\vspace{-1mm}
\label{sec:video_model}

Applying the image-based model developed in Sec.~\ref{sec:image_model} independently to each frame of an unseen video usually results in inconsistent mesh reconstruction (Fig.~\ref{fig:video_quali}(a)), mainly due to the domain differences in video quality, lighting conditions, etc. In this section, we propose to perform online adaptation to fit the model to individual test video, which contains a single object instance that moves over time.
Inspired by the keypoint re-projection constraint described in Sec.~\ref{sec:image_model}, we resort to the UV space, where the (i) RGB texture, and (ii) object parts of an instance should be constant when mapped from 2D via the predicted texture flow, and invariant to shape deformation.
By enforcing the predicted values for (i) and (ii) to be consistent in the UV space across different frames, the adapted network is regularized to generate coherent reconstructions over time.
%
In the following, we describe how to exploit the aforementioned temporal invariances as self-supervisory signals to tune the model.

\paragraph{Part correspondence constraint.}
We propose a part correspondence constraint that utilizes corresponding parts of each video frame to facilitate camera pose learning.
The idea bears resemblance to NR-SFM methods~\cite{novotny2019c3dpo,kong2019deep}, but in contrast, we do not know the ground truth correspondence between frames.
Instead, we resort to an unsupervised video correspondence (UVC) method~\cite{uvc_2019}.
%
The UVC model learns an affinity matrix that captures pixel-level correspondences among video frames. It can be used to propagate any annotation (e.g, segmentation labels, keypoints, part labels, etc.), from an annotated keyframe to the other unannotated frames.
In this work, we generate part correspondence within a clip:
we "paint" a group of random parts on the object, e.g., the vertical stripes in Fig.~\ref{fig:framework}(f), on the first frame and propagate them to the rest of the video using the UVC model.
%

Given the propagated part correspondences in all the frames, we map them to the UV space via the texture flow, similar to our approach for the canonical keypoint map ( Sec.~\ref{sec:image_model}). 
We then average all part UV maps to obtain a video-level part UV map (``UV parts'' in Fig.~\ref{fig:teaser}(a)) for the object depicted in the video. 
%
%
We map the part UV map to each individually reconstructed mesh, and render it via the predicted camera pose of each frame (see Fig.~\ref{fig:teaser}(a), bottom). 
Finally, we penalize the discrepancy between the parts being rendered back to the 2D space, and the propagated part maps, for each frame. 
As the propagated part maps are usually temporally smooth and continuous, this loss implicitly regularizes the network to predict coherent camera pose and shape over time.
In practice, instead of minimizing the discrepancy between the rendered part map and the propagation part map of a frame, we found that it is more robust to penalize the geometric distance between the projections of vertices assigned to each part with 2D points sampled from the corresponding part as:
\vspace{-2mm}
\begin{equation}
L_{c} = \sum_{j=1}^{N_f}{\sum_{i=1}^{N_p} \frac{1}{|V_i^j|}\mathrm{Chamfer}(\mathcal{R}(V_i^j, \theta^j), Y_i^j)},
\label{eq:invar_part}
\end{equation}
where $N_f$ is the number of frames in the video, $N_p=6$ is the number of parts and $V_i^j$ are vertices assigned to part $i$. Here we utilize the $\mathrm{Chamfer}$ distance because the vertex projections $\mathcal{R}(V_i^j, \theta^j)$ do not strictly correspond one-to-one to the sampled 2D points ${Y_i^j}$.


\paragraph{Texture invariance constraint.}
Based on the observation that object texture mapped to the UV space should be invariant to shape deformation and stay constant over time, we propose a texture invariance constraint to encourage consistent texture reconstruction from all frames.
However, naively aggregating the UV texture maps from all the frames via a scheme similar to the one described for keypoints and parts, leads to a blurry video-level texture map.
We instead enforce texture consistency between random pairs of frames, via a swap loss.
Given two randomly sampled frames $I^i$ and $I^j$, we swap their texture maps $I_\mathrm{uv}^\mathbf{i}$ and $I_\mathrm{uv}^\mathbf{j}$, and combine them with the original mesh reconstructions $V^i$ and $V^j$ as:
\vspace{-1mm}
\begin{equation}
L_{t} = \mathrm{dist}(\mathcal{R}(V^i, \theta^i, I_\mathrm{uv}^\mathbf{j})\odot{S^i}, I^i\odot{S^i}) + \mathrm{dist}(\mathcal{R}(V^j, \theta^j, I_\mathrm{uv}^\mathbf{i})\odot{S^j}, I^j\odot{S^j}),
\label{eq:invar_tex}
\end{equation}
where $S^i$ and $S^j$ are the silhouettes of frame $i$ and $j$, respectively and $\mathrm{dist}(\cdot,\cdot)$ is the perceptual metric used in~\cite{zhang2018perceptual,cmrKanazawa18,umr2020}.

\paragraph{Base shape invariance constraint.} 
As discussed in Sec.~\ref{sec:image_model}, our shape model is represented by a base shape $V_\mathrm{base}$ and a deformation term $\Delta V$, in which the base shape $V_\mathrm{base}$ intuitively corresponds to the ``identity'' of the instance, e.g., a duck, or a flying bird, etc.
During online adaptation, we enforce the network to predict consistent $V_\mathrm{base}$ to preserve the identity, via a swapping loss function:
\begin{equation}
L_{s} = \mathrm{niou}(\mathcal{R}(V_\mathrm{base}^\mathbf{j} + \Delta{V^i}, \theta^i), S^i) + \mathrm{niou}(\mathcal{R}(V_\mathrm{base}^\mathbf{i} + \Delta{V^j}, \theta^j),S^j),
\label{eq:invar_base}
\end{equation}
where $V_\mathrm{base}^\mathbf{i}$ and $V_\mathrm{base}^\mathbf{j}$ are the base shapes for frame $i$ and $j$; $\Delta{V^i}$ and $\Delta{V^j}$ are the motion deformations for frame $i$ and $j$; and $\mathrm{niou}(\cdot,\cdot)$ denotes the negative intersection over union (IoU) objective~\cite{kato2019vpl,umr2020}. All other notations are defined in Eq.~\ref{eq:invar_tex}.

\paragraph{As-rigid-as-possible (ARAP) constraint.} Besides the consistency constraints, we keep the ARAP objective, as discussed in Sec.~\ref{sec:image_model}, during online adaptation since it also does not require any form of supervision. We found that the ARAP constraint can obviously improve the qualitative results, as visualized for the online adaptation procedure in Fig.~\ref{fig:no_arap}.

\paragraph{Online adaption.}
During inference, we fine-tune the model on a particular video with the invariance constraints discussed above, along with a silhouette and a texture objective, a Laplacian term as in~\cite{cmrKanazawa18,umr2020}, and the ARAP constraint discussed in Sec.~\ref{sec:image_model}. The foreground masks used for the silhouette and texture objective are obtained by a segmentation model~\cite{chen2017deeplab} trained with the ground truth foreground masks available for the image collection. More details of the objectives used for online adaptation can be found in the supplementary material.

To obtain accurate part propagation of object parts by the UVC~\cite{uvc_2019} model, we employ two strategies. Firstly, we fine-tune all parameters in the reconstruction model on sliding windows instead of all video frames.
Each sliding window includes $N_w=50$ consecutive frames and the sliding stride is set to $N_s=10$. 
We tune the reconstruction model for $N_t=40$ iterations with frames in each sliding window.
Secondly, instead of "painting" random parts onto the first frame and propagating them to the rest of the frames sequentially in a window, we "paint" random parts onto the \textit{middle} frame (i.e. the $\frac{N_w}{2}$th frame) in the window and propagate the parts backward to the first frame as well as forward to the last frame in the window. This strategy improves the propagation quality by decreasing the propagation range to half of the window size.

\vspace{-2mm}
\subsection{Self-supervised Setting}\vspace{-1mm}
\label{sec:unsupervised}
Our model can also be easily generalized to a self-supervised setting in which keypoints are not provided for the image datasets.
In this setting, the template prior as well as camera poses in the CMR method~\cite{cmrKanazawa18} computed from the keypoints are no longer available. 
This self-supervised setting is trained differently from the weakly-supervised one in the following: 
(i) The first stage still assumes shape symmetry to ensure stability when training without keypoints. (ii) It learns a single template from scratch via the progressive training in~\cite{umr2020}. 
(iii) We train this model without the keypoints re-projection and the ARAP constraints in Sec.~\ref{sec:image_model}. (iv) Without the shape bases, the base shape invariance constraint is thus removed in the online adaptation procedure. Other structure and training settings in the self-supervised model are the same as in the weakly-supervised model discussed in Sec.~\ref{sec:image_model} and~\ref{sec:video_model}.

\vspace{-0.10in}
\section{Experiments}
\vspace{-0.10in}
We conduct experiments on animals, i.e., birds and zebras. We evaluate our contributions in two aspects: (i) the improvement of single-view mesh reconstruction, and (ii) the reconstruction of a sequence of frames via online adaptation. 
Due to the lack of ground truth meshes for images and videos captured in the wild, we evaluate the reconstruction results via mask and keypoint re-projection accuracy, e.g., we follow, and compare against \cite{cmrKanazawa18} to evaluate the model trained on the image dataset. We also describe a new bird video dataset that we curate and evaluate the test-time tuned model on it in the following.
We focus on evaluations on the bird category in the paper and leave evaluations on the zebra category to the supplementary.

\begin{figure}
  \centering
    \includegraphics[width=0.9\textwidth]{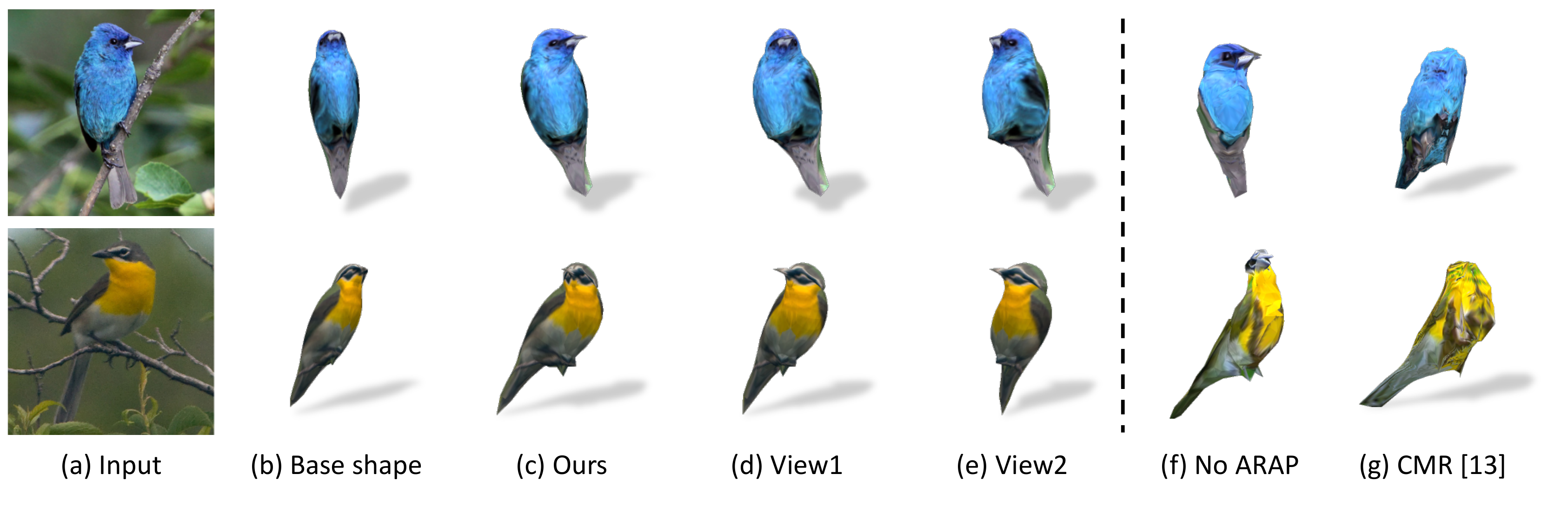}
  \caption{\footnotesize Mesh reconstructions from single-view images. All meshes are visualized from the predicted camera pose except for (d) and (e), where the reconstructions in (c) are visualized from two extra views. Meshes in (f) are reconstructed by a model trained without the ARAP constraint.}\label{fig:img_quali}\vspace{-5mm}
\end{figure}

\vspace{-2mm}
\subsection{Experimental Settings}\vspace{-1mm}
\label{sec:settings}
\paragraph{Datasets.} 
We first train image reconstruction models, discussed in Sec.~\ref{sec:image_model}, for the CUB bird~\cite{wah2011caltech} and the synthetic zebra~\cite{Zuffi:ICCV:2019} datasets.
For test-time adaptation on videos, we collect a new bird video dataset for quantitative evaluation.
Specifically, we collect 19 slow-motion, high-resolution bird videos from YoutTube, and 3 bird videos from the DAVIS dataset~\cite{KhoRohrSch_ACCV2018}. 
For each slow-motion video collected from the Internet, we apply a segmentation model~\cite{chen2017deeplab} trained on the CUB bird dataset~\cite{wah2011caltech} to obtain its foreground segmentation for online adaptation.

\paragraph{Evaluation metrics.} 
We evaluate the image-based model on the testing split of the CUB dataset.
Note that for keypoint re-projection, instead of using the keypoint assignment matrix in \cite{cmrKanazawa18}, we apply the canonical keypoint UV map to obtain the 3D keypoints (Sec. \ref{sec:image_model}).
For the video dataset, we annotate frame-level object masks and keypoints via a semi-automatic procedure. We train a segmentation model and a keypoint detector~\cite{guo2019aligned} on the CUB dataset. Then, we manually adjust and filter out inaccurate predictions to ensure the correctness of the ground-truth labels.
To evaluate the accuracy of mask re-projection, we compute the Jacaard index $\mathcal{J}$ (IoU) and contour-based accuracy $\mathcal{F}$ proposed in~\cite{pont20172017}, between the rendered masks and the ground truth silhouettes of all annotated frames.
Evaluations on keypoint re-projection can be found in the supplementary documents.

In addition, to further quantitatively evaluate shape reconstruction quality, we animate a synthetic 3D bird model and create a video with 520 frames in various poses such as flying, landing, walking etc., as shown in Fig.~\ref{fig:quali_3D}.
We then compare the predicted mesh with the ground truth mesh using Chamfer distance every 10 frames.
%
\paragraph{Network architecture.} For fair comparisons to the baseline method~\cite{cmrKanazawa18}, we train our model using the same network as~\cite{cmrKanazawa18}, i.e., ResNet18~\cite{he2016deep} with batch normalization layers~\cite{ioffe2015batch} as the encoder, we call this model ``ACMR'' in the following, which is short for ``asymmetric CMR''. However, ACMR cannot be well adapted to test videos due to the batch normalization layers and the domain gap between images and videos (see Table~\ref{tab:quanti_video}(d)). Thus, we train a variant model where we use ResNet50~\cite{he2016deep} as our encoder and replace all the batch normalization layers in the network with group normalization layers~\cite{wu2018group}. We call this variant model ``ACMR-vid''. All test-time training is carried out on the ACMR-vid model unless otherwise specified.

\paragraph{Network training.}
Taking the weakly-supervised setting as an example, 
to train the image reconstruction model, we first warm up the model without the motion deformation branch, the keypoint re-projection objective, or the ARAP constraint for $200$ epochs. 
This warm-up process effectively avoids the trivial solution where the model solely depends on the motion deformation branch for shape deformation while ignoring the base shape branch.
We then train the full image reconstruction network with all objectives for another $200$ epochs.
%
%
%
%
Other training details, including the self-supervised setting, and the illustration of our network architecture can be found in the supplementary material.

\vspace{-2mm}
\subsection{Qualitative Results}
\vspace{-1mm}
\paragraph{Mesh reconstruction from images.}
In Fig.~\ref{fig:img_quali}, we show visualizations of reconstructed bird meshes from single-view images. 
Thanks to the ``motion deformation branch'' discussed in Sec.~\ref{sec:image_model}, the proposed ACMR model is able to capture asymmetric motion of the bird such as head rotation (Fig.~\ref{fig:img_quali}(c)), which cannot be modeled by the baseline method~\cite{cmrKanazawa18} (Fig.~\ref{fig:img_quali}(g)).

\begin{figure}
  \centering
    \includegraphics[width=0.95\textwidth]{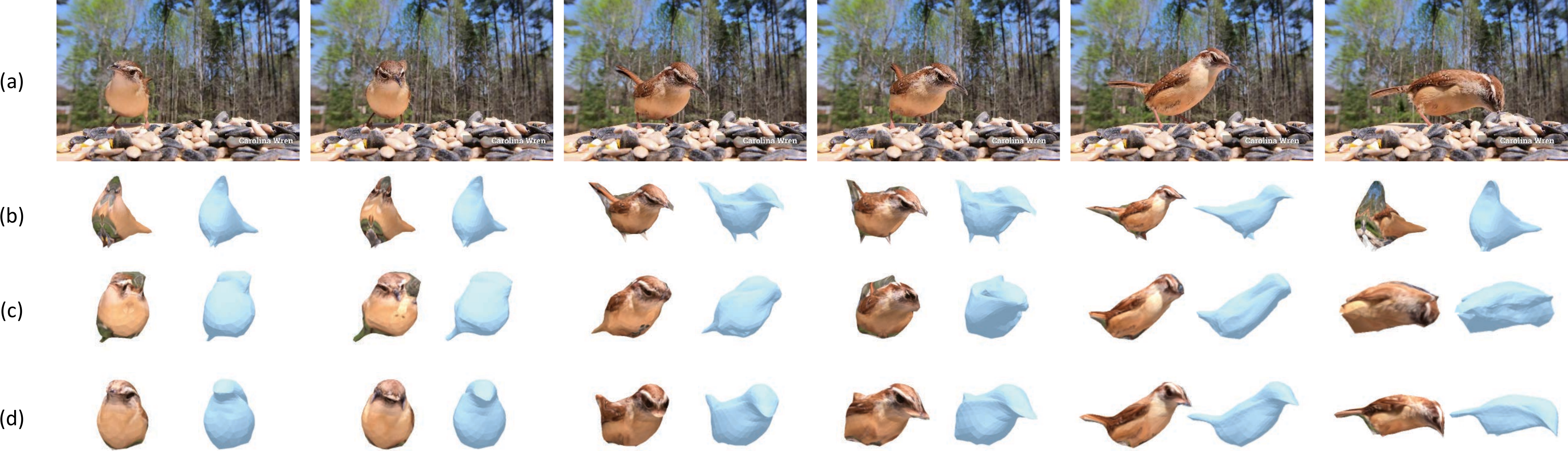}
  \caption{\footnotesize Mesh reconstruction from video frames. (a) Input video frames. (b) Reconstruction from the model trained only on single-view images. (c) Reconstruction from the model test-time trained on the video without the invariance constraints in Sec.~\ref{sec:video_model}. (d) Reconstruction from the proposed video reconstruction model.}\label{fig:video_quali}\vspace{-3mm}
\end{figure}

\begin{figure}[h]
  \centering
    \includegraphics[width=0.9\textwidth]{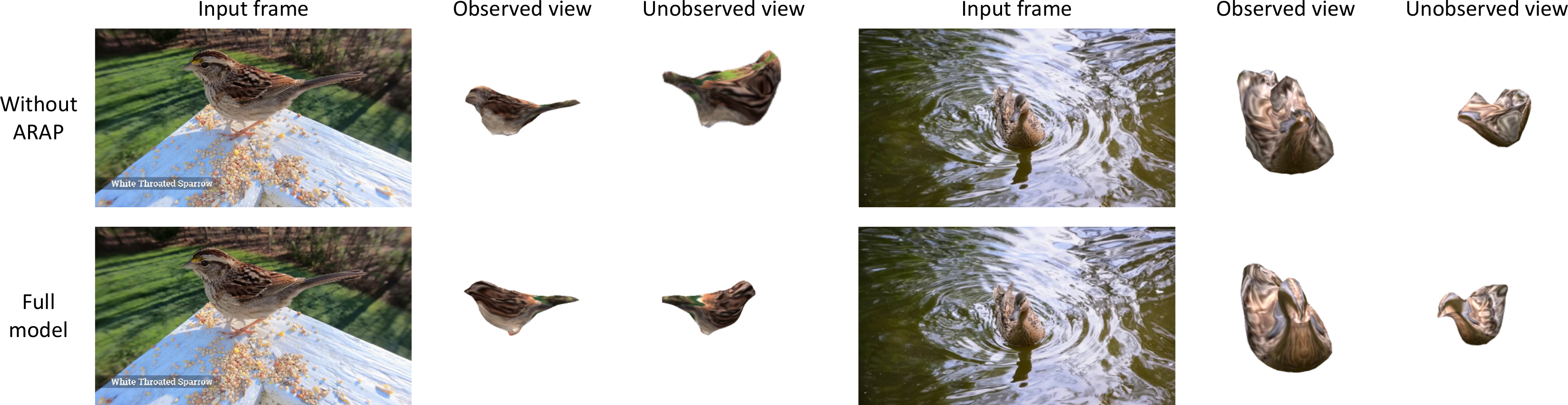}
  \caption{\footnotesize Qualitative comparison of online adaptation with/without the ARAP constraint.}\label{fig:no_arap}
  \vspace{-7pt}
\end{figure}

\paragraph{Online adaptation on a video.}
We visualize the reconstructed meshes by our ACMR-vid model for video frames in Fig.~\ref{fig:video_quali}. Without online adaptation, the ACMR-vid model independently applied to each frame suffers from a domain gap and shows instability over time (Fig.~\ref{fig:video_quali}(b)). 
With online adaption as discussed in Sec.~\ref{sec:video_model}, the ACMR-vid model reconstructs plausible meshes for each video frame as shown in Fig.~\ref{fig:video_quali}(c) and (d).
Specifically, to demonstrate the effectiveness of the proposed invariance constraints, we also show reconstructions of an ACMR-vid model trained without all the invariance constraints in Fig.~\ref{fig:video_quali}(c), which predicts less reliable shape, camera pose as well as texture compared to our full ACMR-vid model.
Finally, we visualize the effectiveness of ARAP for online adaptation in Fig.~\ref{fig:no_arap}. Without this constraint, the reconstructed meshes are less plausible, especially from unobserved views.
More video examples can be found in the supplementary.

\vspace{-2mm}
\subsection{Quantitative Results}\vspace{-1mm}

\paragraph{Evaluations on the image dataset.}
\begin{table}[h]
\centering
  \caption{\small{Quantitative evaluation of mask IoU and keypoint re-projection (PCK@0.1) on the CUB dataset~\cite{wah2011caltech}. }} 
  \label{tab:quanti}
  {\footnotesize{
  \begin{tabular}{c|c|c|c|c|c}
  \hline
  (a) Metric & (b) CMR~\cite{cmrKanazawa18} & (c) ACMR &  \specialcell{(d) ACMR,\\no $\Delta{V}$} &  \specialcell{(e) ACMR,\\no ARAP} & (f) ACMR-vid\\
  \hline
  Mask IoU~$\uparrow$ & 0.706 & 0.708 & 0.647 & 0.758 & \textbf{0.773}\\
  PCK@0.1 ~$\uparrow$ &0.810 & 0.855 & 0.790 & 0.857 & \textbf{0.895}\\
  \hline 		 
  \end{tabular}
  }}
  \vspace{-2mm}
\end{table}

As shown in Table~\ref{tab:quanti}(b) \textit{vs.}\ (c), our ACMR model achieves comparable mask IoU and higher keypoints re-projection accuracy compared to the baseline model~\cite{cmrKanazawa18} with the same network architecture. This confirms the correctness of both the reconstructed meshes as well as that of the predicted camera poses.
In addition, our ACMR-vid model achieves even better performance as shown in Table~\ref{tab:quanti}(f). 
We note that our full ACMR model does not quantitatively outperform the model trained without the ARAP constraint, because the motion deformation $\Delta V$ freely over-fits to the mask and keypoint supervision without any regularization. However, the model without the ARAP constraint visually shows spikes and unnatural deformations as shown in Fig.~\ref{fig:img_quali}(f) and in the supplementary. 
%

%
\begin{wrapfigure}{r}{0.5\textwidth}
  \begin{centering}
    \includegraphics[width=0.45\textwidth]{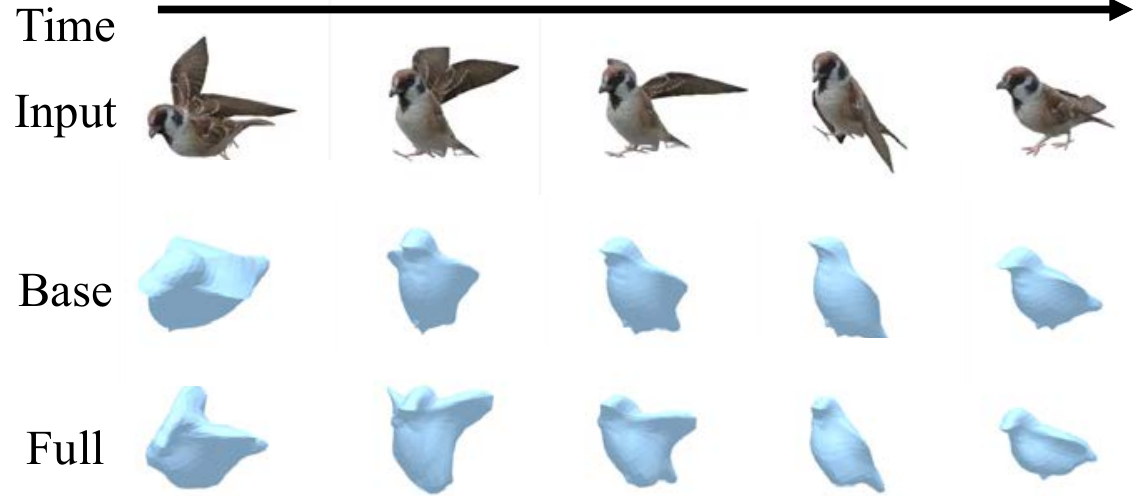}
  \end{centering}
  \caption{\footnotesize  Reconstructions of an animated video clip.\vspace{-10pt}}\label{fig:quali_3D}
  \vspace{-5pt}
\end{wrapfigure}

\paragraph{Evaluations on the video dataset.}

%
%
%
%
%
\begin{table}[h]
\vspace{-3mm}
\centering
  \caption{\small{Quantitative evaluation of mask re-projection accuracy on the bird video dataset. ``(T)'' indicates the model is test-time trained on the given video., $L_c$, $L_t$, $L_s$ are defined in Eq.~\ref{eq:invar_part},~\ref{eq:invar_tex},~\ref{eq:invar_base} respectively.}}
  \label{tab:quanti_video}
  
  {\footnotesize{
  \begin{tabular}{c|c|c|c|c|c}
  \hline
  (a) Metric & (b) CMR~\cite{cmrKanazawa18} & (c) ACMR & (d) ACMR (T) & \specialcell{(e) ACMR-vid (T),\\no $L_c$, $L_t$, $L_s$} &  (f) ACMR-vid (T)\\
  \hline
  $\mathcal{J} (Mean)\uparrow$ & 0.554 & 0.686 & 0.706 & 0.836 & \textbf{0.868}\\
  $\mathcal{F} (Mean)\uparrow$ & 0.209 & 0.363 & 0.406 & 0.666 & \textbf{0.756}\\
  \hline 		 
  \end{tabular}
  }}
  \vspace{-2mm}
\end{table}
As shown in Table~\ref{tab:quanti_video}(b) and (c) \textit{vs.}\ (e), by using the proposed online adaptation method discussed in Sec.~\ref{sec:video_model}, the model tuned on videos achieves higher $\mathcal{J}$ and $\mathcal{F}$ scores compared to the model trained only on images. This indicates that the test-time trained model successfully adapts to unlabeled videos and can reconstruct meshes that conform well to the frames.
The performance of the model is further improved by adding the correspondence, texture, and shape invariance constraints discussed in Sec.~\ref{sec:video_model} during online adaptation, as shown in Table~\ref{tab:quanti_video}(f).

\textbf{Evaluation on animated sequences.} We apply the proposed model to an animated video clip and compare the predicted mesh with the ground truth mesh using Chamfer distance every 10 frames. We show the qualitative reconstructions in Fig.~\ref{fig:quali_3D} and quantitative evaluation results in Table~\ref{tab:quanti_3D}. 
The proposed ACMR method outperforms the baseline CMR~\cite{cmrKanazawa18} model and is further improved via the proposed online adaptation strategy discussed in Sec.~\ref{sec:video_model}.

\vspace{-2mm}
\begin{wraptable}{r}{6.6cm}
\vspace{-3mm}
\begin{tabular}{c|c|c|c}
\hline
\footnotesize{Metric}& \footnotesize{CMR} & \footnotesize{ACMR} & \footnotesize{ACMR (T)}\\
\hline
\footnotesize{Chamfer\textbf{$\downarrow$}}& \footnotesize{0.016} & \footnotesize{0.015}
&\footnotesize{\textbf{0.012}}\\
\hline
\end{tabular}\vspace{-2mm}
\caption{\scriptsize{Evaluation on synthetic data.}}
\label{tab:quanti_3D}
\vspace{-4mm}
\end{wraptable}

\paragraph{Evaluations for self-supervised setting (Sec.~\ref{sec:unsupervised}).} After online adaptation, this model too, achieves both a higher $\mathcal{J}$ score (0.843 \textit{vs.} 0.582) and $\mathcal{F}$ score (0.678 \textit{vs.} 0.256) compared to the model pre-trained on the image dataset, i.e., our method reconstructs promising dynamic 3D meshes without annotating any keypoint for training.
\vspace{-0.10in}
\section{Conclusions}
\vspace{-0.10in}
We propose a method to reconstruct temporally consistent 3D meshes of deformable objects from videos captured in the wild. We learn a category-specific 3D mesh reconstruction model that jointly predicts the shape, texture, and camera pose from single-view images, which is capable of capturing asymmetric non-rigid motion deformation of objects. We then adapt this model to any unlabeled video by exploiting self-supervised signals in videos, including those of shape, texture, and part consistency. Experimental results demonstrate the superiority of the proposed method compared to state-of-the-art works, both qualitatively and quantitatively.

\section*{Broader Impact}
The developed method will make significant contributions to both the 3D vision and endangered species research. The method provides a way to study animals that can only be captured in the wild as 2D videos, e.g., endangered animal species of birds and zebras. The broader impact includes enhancing our understanding of such endangered animals simply from videos, as they can be reconstructed and viewed in 3D. The method can also be applied to tasks such as bird watching, motion analysis, shape analysis, to name a few. Furthermore, another important application is to simplify an artists workflow, as an initial animated and textured 3D shape can be directly derived from a video.

\clearpage
\section*{\Large Appendix}
In the Appendix, we provide additional details, discussions, and experiments to support the original submission.
In the following, we first discuss our self-supervised setting in Sec.~\ref{asec:unsupervised}. We then describe evaluation of keypoint re-projection accuracy on videos in Sec.~\ref{asec:kp_proj}. Next, we show more ablation studies in Sec.~\ref{asec:ablation}. More qualitative results on both bird and zebra image reconstructions are present in Sec.~\ref{asec:quali}. Details of the network design and implementation are discussed in Sec.~\ref{asec:net_arch} and Sec.~\ref{asec:training_details}, respectively. Finally, we describe failure cases and limitations in Sec.~\ref{asec:failure_cases}.

\vspace{-2mm}
\section{Self-supervised Mesh Reconstruction}
\begin{figure}[h]
  \centering
    \includegraphics[width=0.9\textwidth]{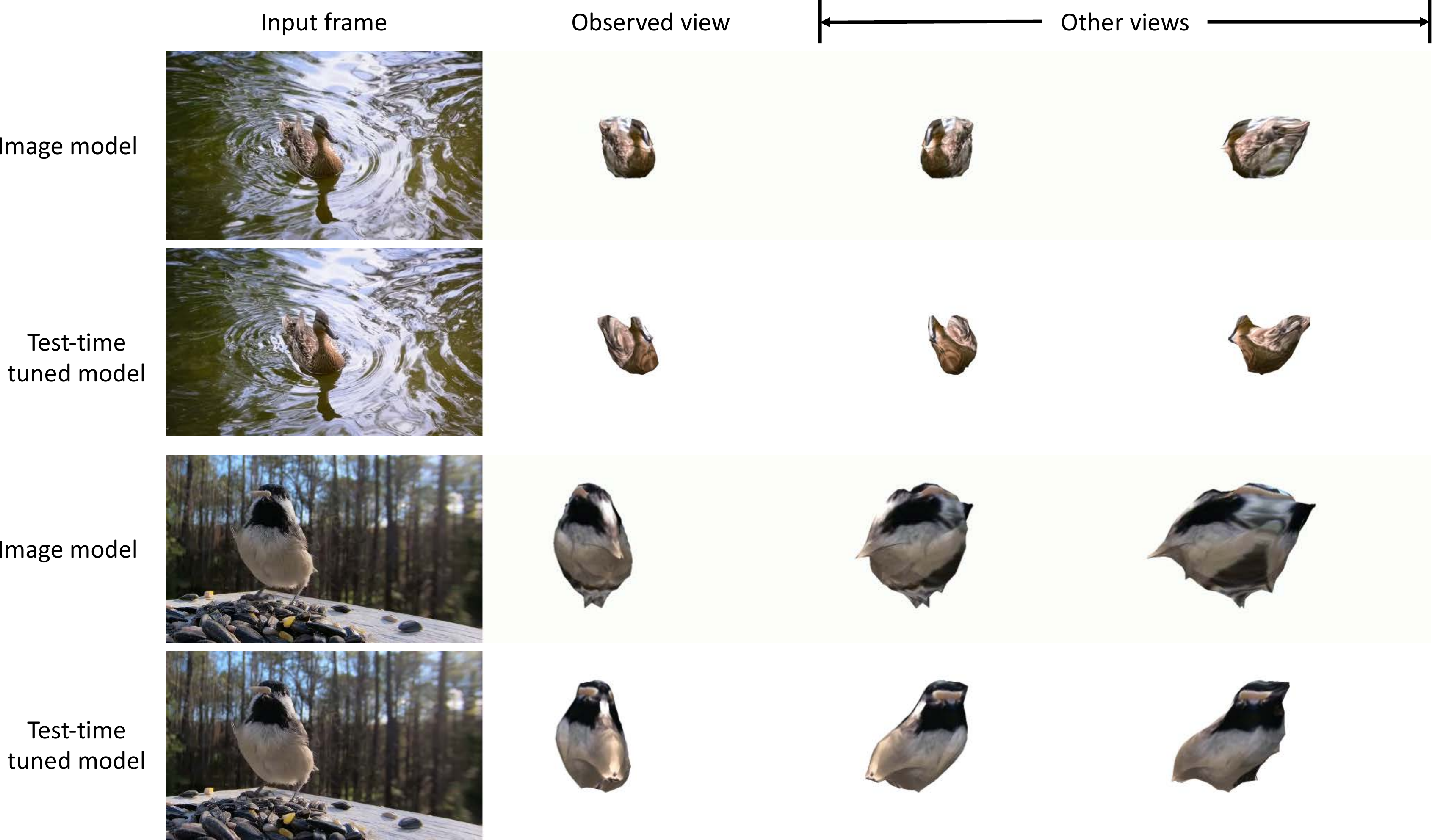}
  \caption{\footnotesize Comparison of the self-supervised image model and the test-time tuned self-supervised model.}\label{fig:unsup}
  \vspace{-7pt}
\end{figure}

\vspace{-1mm}
\label{asec:unsupervised}
We train the self-supervised image reconstruction model with only silhouettes and single-view images in a category. To this end, we also learn a template from scratch as in the self-supervised 3D mesh reconstruction approach~\cite{umr2020}. Essentially, we do not apply the semantic parts from the SCOPS method~\cite{hung:CVPR:2019}, which means that no additional modules are required for training. After training the image model, we adapt it to each unlabeled video using the method discussed in Sec.3.2 in the submission. 
Since neither keypoints annotations, nor additional self-supervised blocks such as SCOPS are adopted, the results of this self-supervised single-view image reconstruction model do not outperform those of existing methods, i.e., ~\cite{cmrKanazawa18} and ~\cite{umr2020}, and of the proposed ACMR model (see Sec. 4.3 for the quantitative results). However, the test-time training improves the fidelity and the robustness of the reconstruction results as shown in Fig.~\ref{fig:unsup}. The reconstructions are more plausible after online adaptation, especially from unobserved views.

\section{Keypoint Re-projection Accuracy on Videos}
\label{asec:kp_proj}
 
\begin{wrapfigure}{r}{0.55\textwidth}
  \vspace{-15pt}
  \centering
    \includegraphics[width=0.5\textwidth]{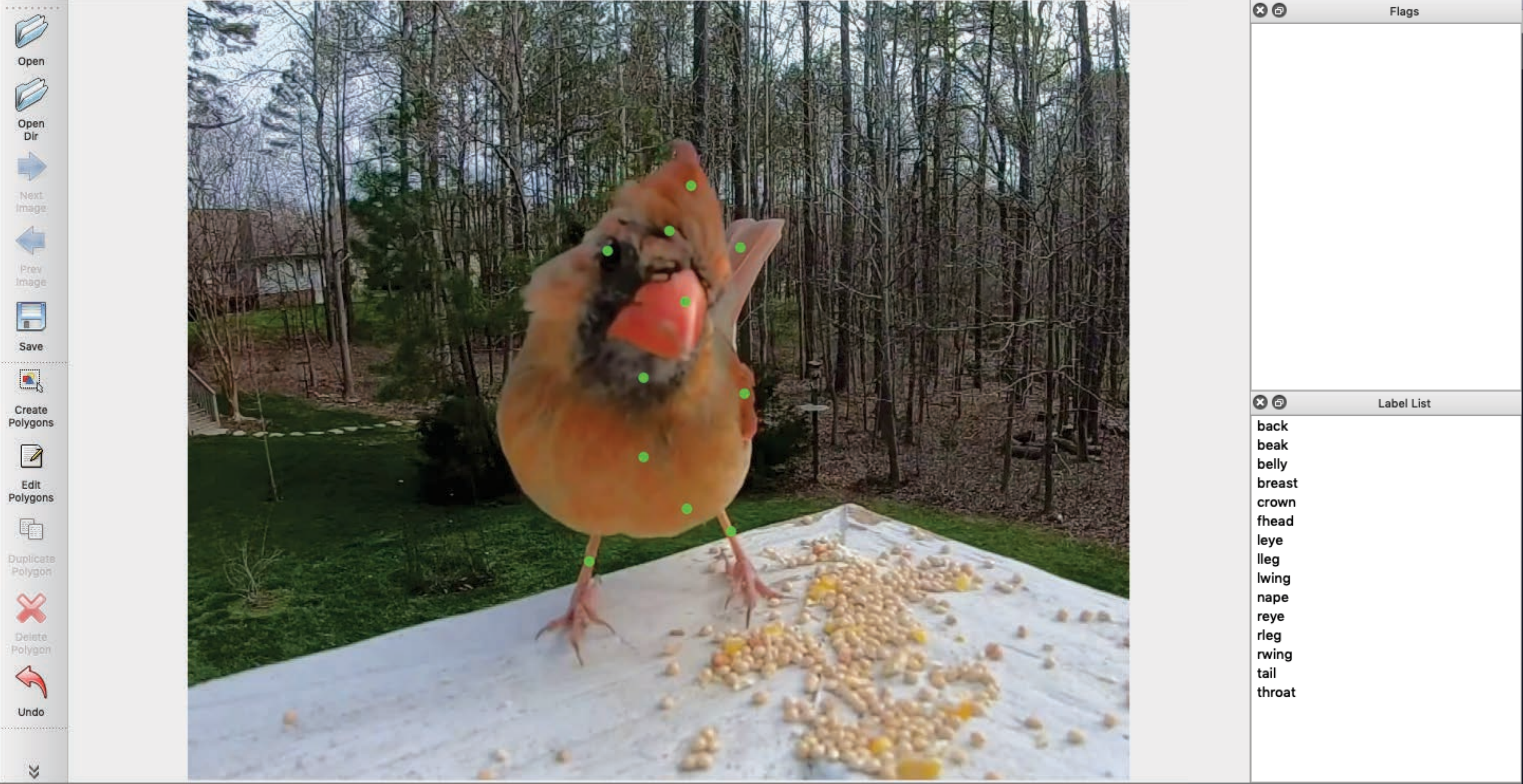}
  \caption{\footnotesize Keypoints annotation using the Labelme~\cite{labelme2016} toolkit.}\label{fig:anno}
  \vspace{-10pt}
\end{wrapfigure}

\begin{figure}[t]
  \centering
    \includegraphics[width=0.98\textwidth]{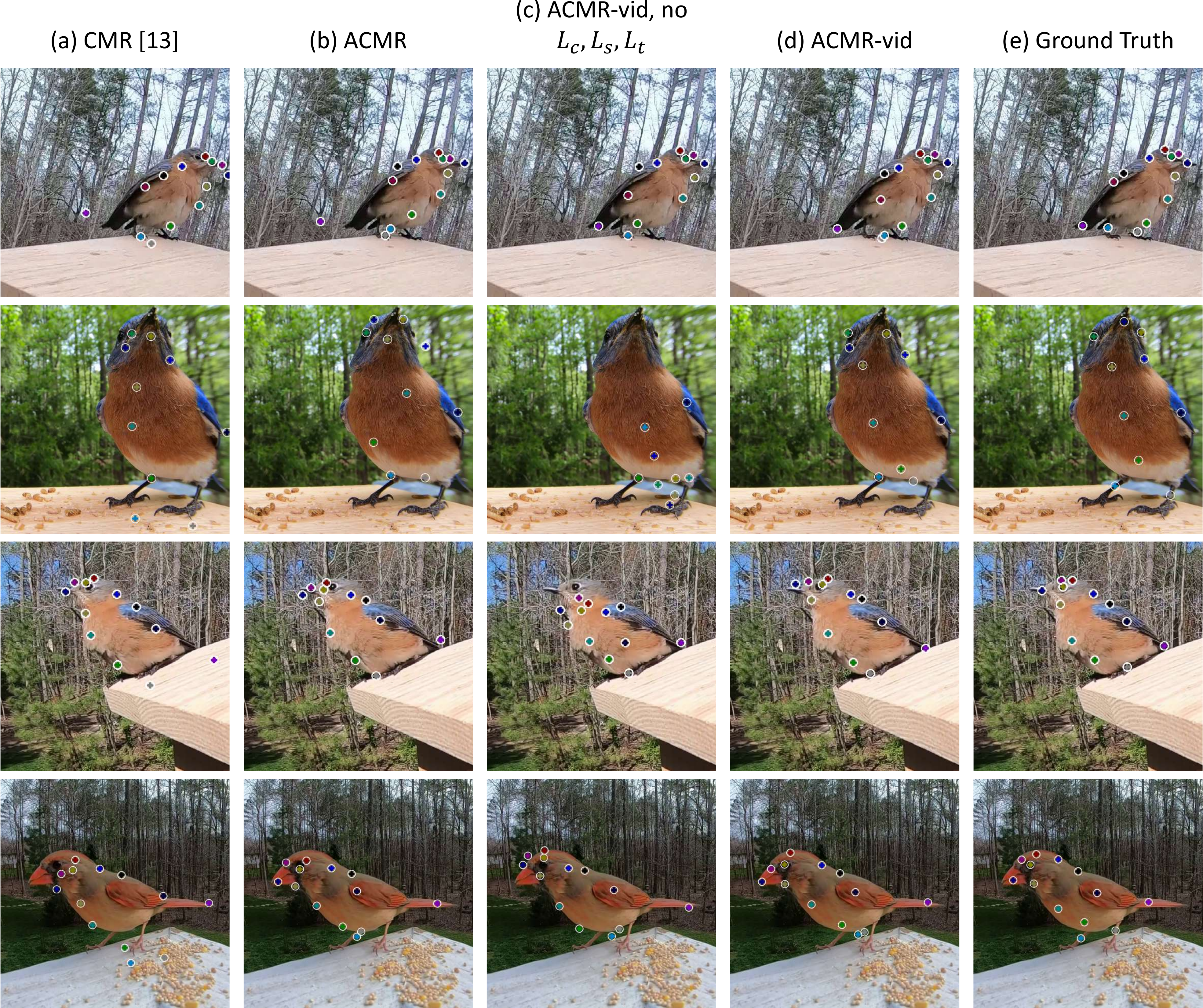}
  \caption{\footnotesize Visualization of re-projected keypoints on videos. We use white circles to highlight the keypoints for better visualization.}\label{fig:video_kp}
  \vspace{-7pt}
\end{figure}

\paragraph{Video Keypoints Annotation.} 
We evaluate the keypoint re-projection accuracy (see Sec 4.1 in the paper) on the 22 videos we collected. To create the ground truth keypoints, we follow the protocol of the CUB dataset~\cite{wah2011caltech} to annotate 15 semantic keypoints every five frames in each video, via the Labelme~\cite{labelme2016} toolkit (see Fig.~\ref{fig:anno} for the annotation interface.)
Visualizations of the re-projected keypoints by different methods are visualized and compared in Fig.~\ref{fig:video_kp}.

\paragraph{Details of Keypoint Re-projection.} Since we do not have the keypoint assignment matrix proposed in \cite{cmrKanazawa18}, we employ the canonical keypoint UV map to obtain the 3D keypoints (Sec. 3.1 in the paper). The keypoint re-projection is done by (i) warping the canonical keypoint UV map to each individual predicted mesh surface; (ii) projecting the canonical keypoint back to the 2D space via the predicted camera pose; (iii) comparing against the ground truth keypoints in 2D. This evaluation implicitly reveals the correctness of both the predicted shape and camera pose for the mesh reconstruction algorithm, especially for objects that do not have 3D ground truth annotations.
 
Compared to frame-wisely applying CMR~\cite{cmrKanazawa18} (Table~\ref{tab:kp_proj_quanti} (b)) or ACMR (Table~\ref{tab:kp_proj_quanti} (c)) discussed in Sec. 3.1 in the main paper, the test-time tuned model achieves higher PCK score, as shown in Table~\ref{tab:kp_proj_quanti} (f). It verifies the effectiveness of the proposed test-time training procedure and the invariance constraints. 
Essentially, as we noted in Sec 4.1, although the original ACMR, i.e., using the ResNet-18~\cite{he2016deep} as the image encoder with batch normalization layers~\cite{ioffe2015batch} in Table~\ref{tab:kp_proj_quanti} (c), achieves relatively promising results, it is hard to adapt this model to new domains like low-quality videos (e.g., when switching from the \textit{.eval()} mode to the \textit{.train()} mode in PyTorch~\cite{NEURIPS2019_9015}). The performance drops significantly after test-time tuning as shown in Table~\ref{tab:kp_proj_quanti} (d).

 \begin{table}[t]
\centering
  \caption{\small{Keypoint re-projection evaluation on videos. ``(T)'' indicates the model is test-time trained on the given video., $L_c$, $L_t$, $L_s$ are defined in Eq.4, 5 and 6 of the main paper, respectively.}}
  \label{tab:kp_proj_quanti}
  {\footnotesize{
  \begin{tabular}{c|c|c|c|c|c}
  \hline
  (a) Metric & (b) CMR~\cite{cmrKanazawa18} & (c) ACMR & (d) ACMR (T) & \specialcell{(e) ACMR-vid (T),\\no $L_c$, $L_t$, $L_s$} &  (f) ACMR-vid (T)\\
  \hline
  $PCK@0.1\uparrow$ & 0.514 & 0.751 & 0.424  & 0.644 & \textbf{0.794} \\
  \hline 		 
  \end{tabular}
  }}
  \vspace{-2mm}
\end{table}

\vspace{-2mm}
\section{Ablation Studies}
\vspace{-1mm}
\label{asec:ablation}

\begin{wraptable}{r}{6.5cm}
  \caption{\small{Quantitative comparison of the single base model with the proposed ACMR model.}}
  \label{tab:ablation}
  {\footnotesize{
  \begin{tabular}{c|c|c}
  \hline
  (a) Metric & (b) Single base & (c) ACMR\\
  \hline
  Mask IoU~$\uparrow$ & 0.605 & 0.708\\
  PCK@0.1 ~$\uparrow$ & 0.655 & 0.855\\
  \hline 		 
  \end{tabular}
  }}
\end{wraptable}

\subsection{The Role of Shape Base} 
\begin{figure}[h]
  \centering
    \includegraphics[width=0.9\textwidth]{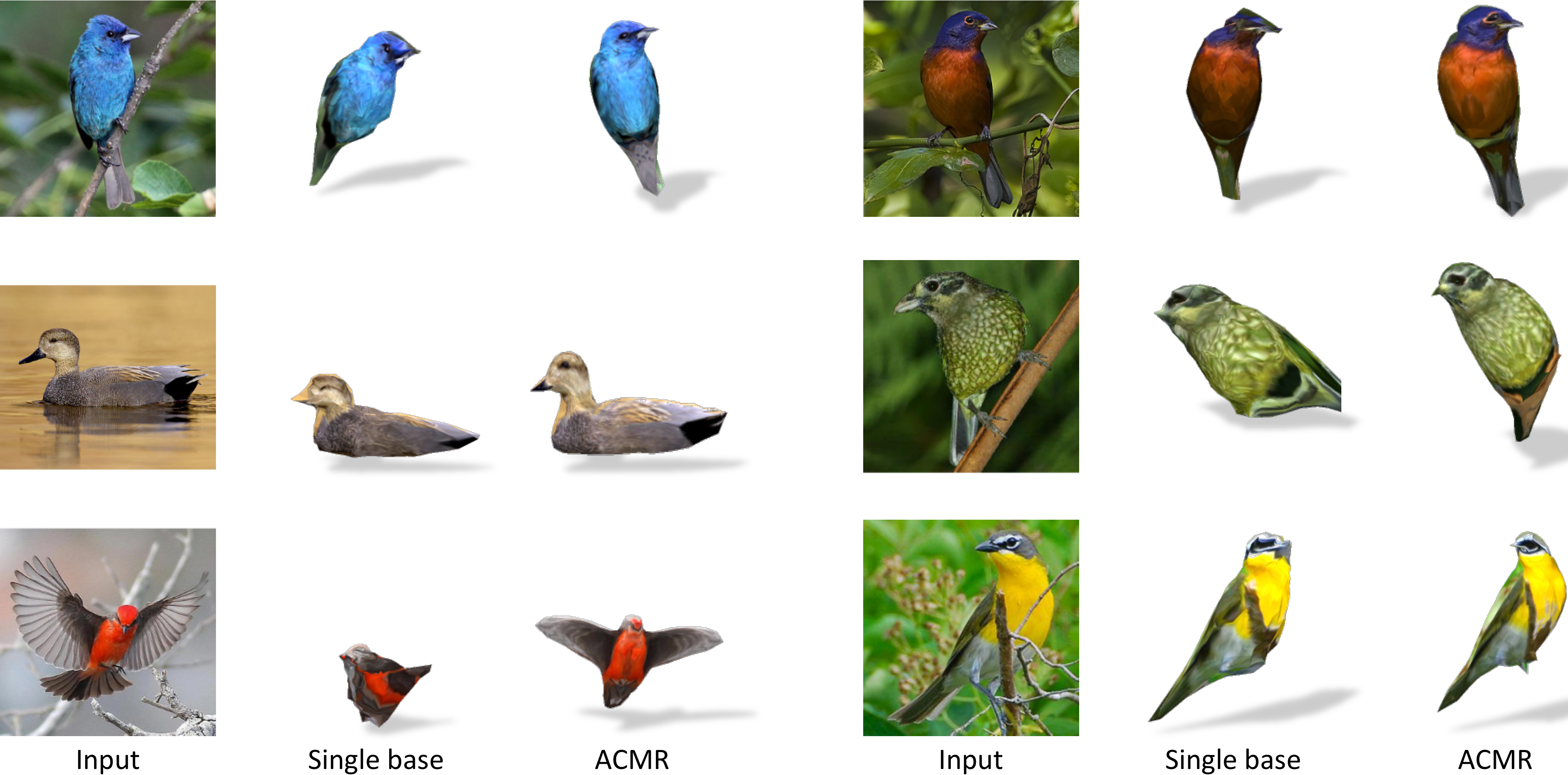}
  \caption{\footnotesize Qualitative comparison of the single base model with the proposed ACMR model with multiple shape bases. The single base model suffers when the instance is largely different from the template, e.g. flying bird or a duck.}\label{fig:ablation}
  \vspace{-7pt}
\end{figure}

To demonstrate the superiority of using a set of shape bases versus using a single template, we train a baseline model where we replace the shape combination branch with the template obtained by the CMR approach~\cite{cmrKanazawa18}. This setting is equivalent to a using a single shape base (denoted as \textit{single base}).
We show quantitative and qualitative comparisons with the proposed ACMR model in Table~\ref{tab:ablation} and Fig.~\ref{fig:ablation}, respectively.
%
As shown in Table~\ref{tab:ablation}(b) \textit{vs.} (c), the model trained with a single base template struggles to fit the final shape when the instance is largely different from the given template (also shown in Fig.~\ref{fig:ablation}). In contrast, the proposed ACMR model with 8 shape bases performs favorably against the single base model.

\subsection{ARAP Constraint in Online Adaptation}
\vspace{-1mm}
\begin{wraptable}{r}{7.5cm}
  \vspace{-3mm}
  \caption{\small{Ablation study on the ARAP constraint in online adaptation.}}
  \label{tab:no_arap}
  {\footnotesize{
  \begin{tabular}{c|c|c}
  \hline
  (a) Metric & (b) without ARAP & (c) ACMR-vid (T)\\
  \hline
  $\mathcal{J} (Mean)\uparrow$ & 0.875 & 0.868\\
  $\mathcal{F} (Mean)\uparrow$ & 0.782 & 0.756\\
  PCK@0.1$\uparrow$ & 0.815 & 0.794\\
  \hline 		 
  \end{tabular}
  }}
\end{wraptable}

To verify the effectiveness of using the ARAP constraint in the online adaptation process, we test-time tune on the videos without this constraint.
Although performing online adaptation without the ARAP constraint yields better quantitative evaluations as shown in Table~\ref{tab:no_arap}, the reconstructed meshes are not plausible from unobserved views, as shown in Fig.6 in the submission.

\vspace{-2mm}
\section{Qualitative Evaluations}
\vspace{-1mm}
\label{asec:quali}
\begin{figure}[h]
  \centering
    \includegraphics[width=0.98\textwidth]{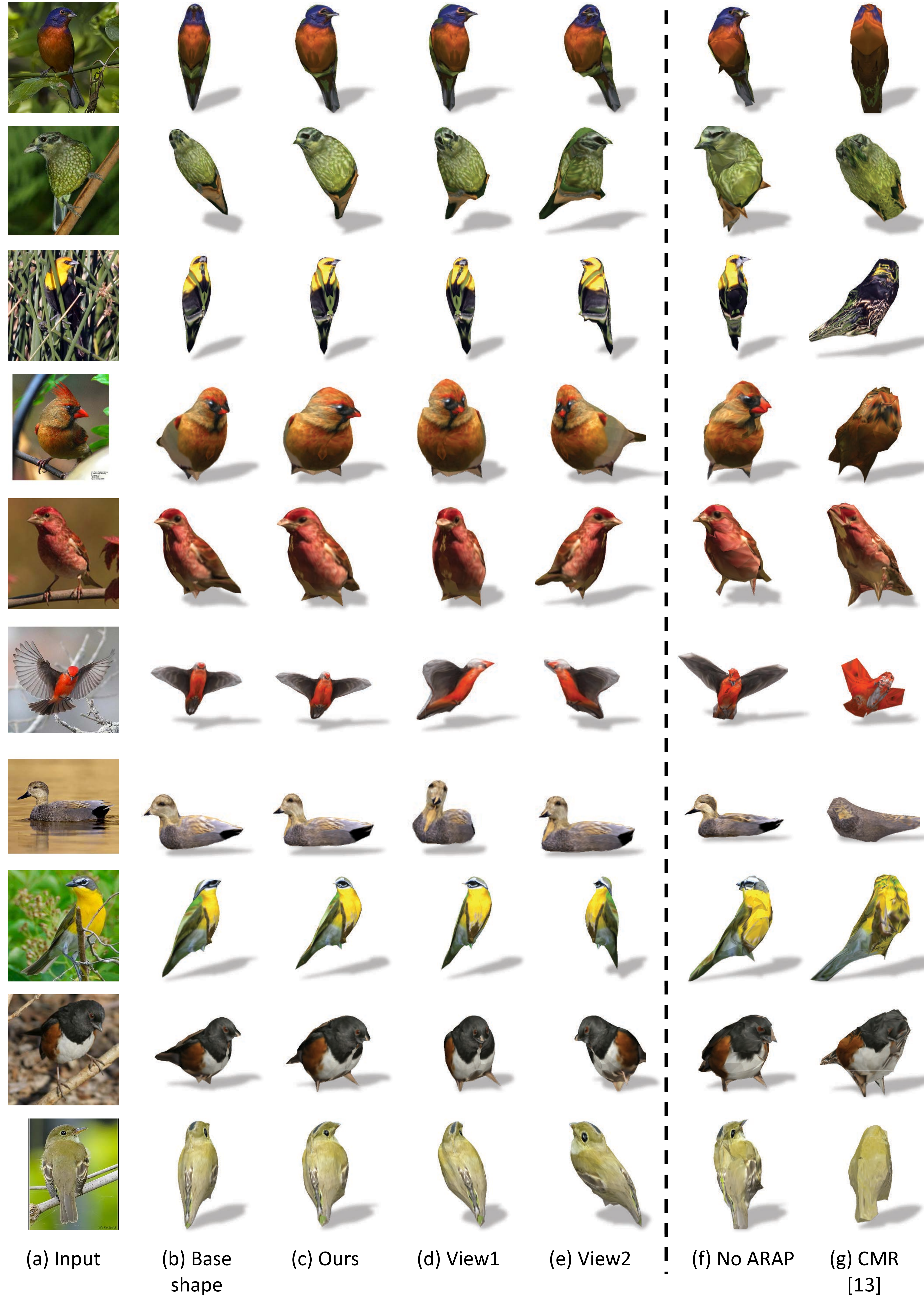}
  \caption{\footnotesize More qualitative reconstruction results on CUB birds~\cite{wah2011caltech}.}\label{fig:image_quali}
  \vspace{-7pt}
\end{figure}

\subsection{Camera Pose Stability}
\begin{figure}[h]
  \centering
    \includegraphics[width=0.98\textwidth]{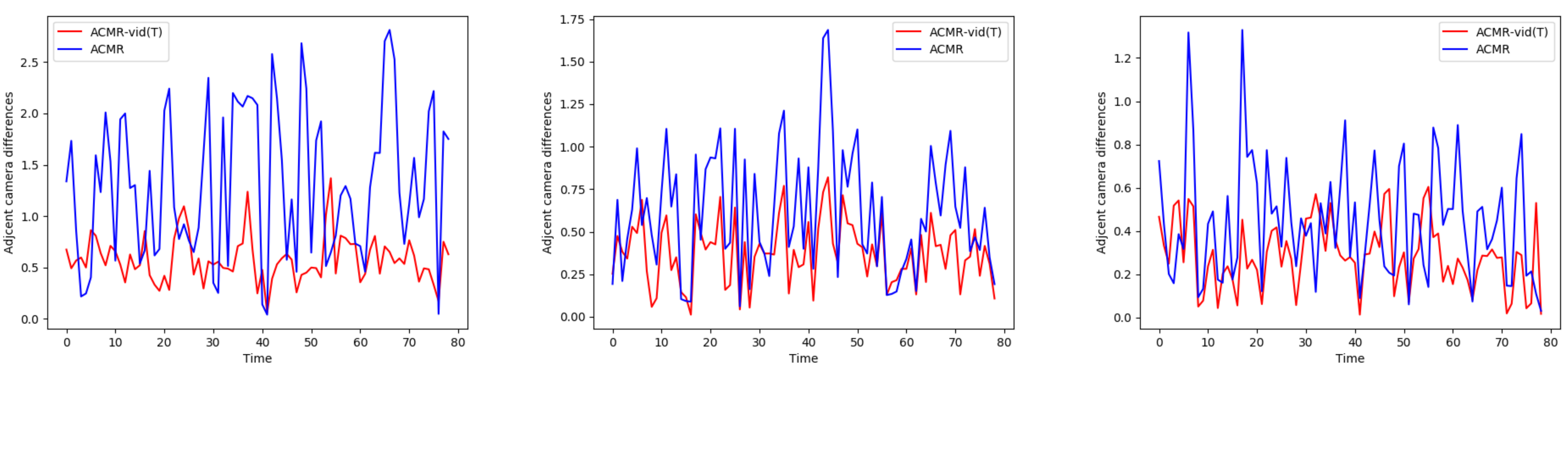}
  \caption{\footnotesize Camera stability visualization. We visualize differences between adjacent camera pose predictions. The blue and red lines represent the model trained only with images and the test-time tuned model, respectively.}\label{fig:cam_vis}
  \vspace{-7pt}
\end{figure}

To visually demonstrate the effectiveness of the test-time training procedure that stabilizes camera pose prediction, we visualize the differences in camera pose predictions between adjacent frames in Fig.~\ref{fig:cam_vis}. Compared to the model that is only trained on images, the proposed method predicts more stable camera poses that change smoothly over time.

\subsection{Keypoints Re-projection for Image-based Reconstruction}
\begin{figure}[h]
  \centering
    \includegraphics[width=0.98\textwidth]{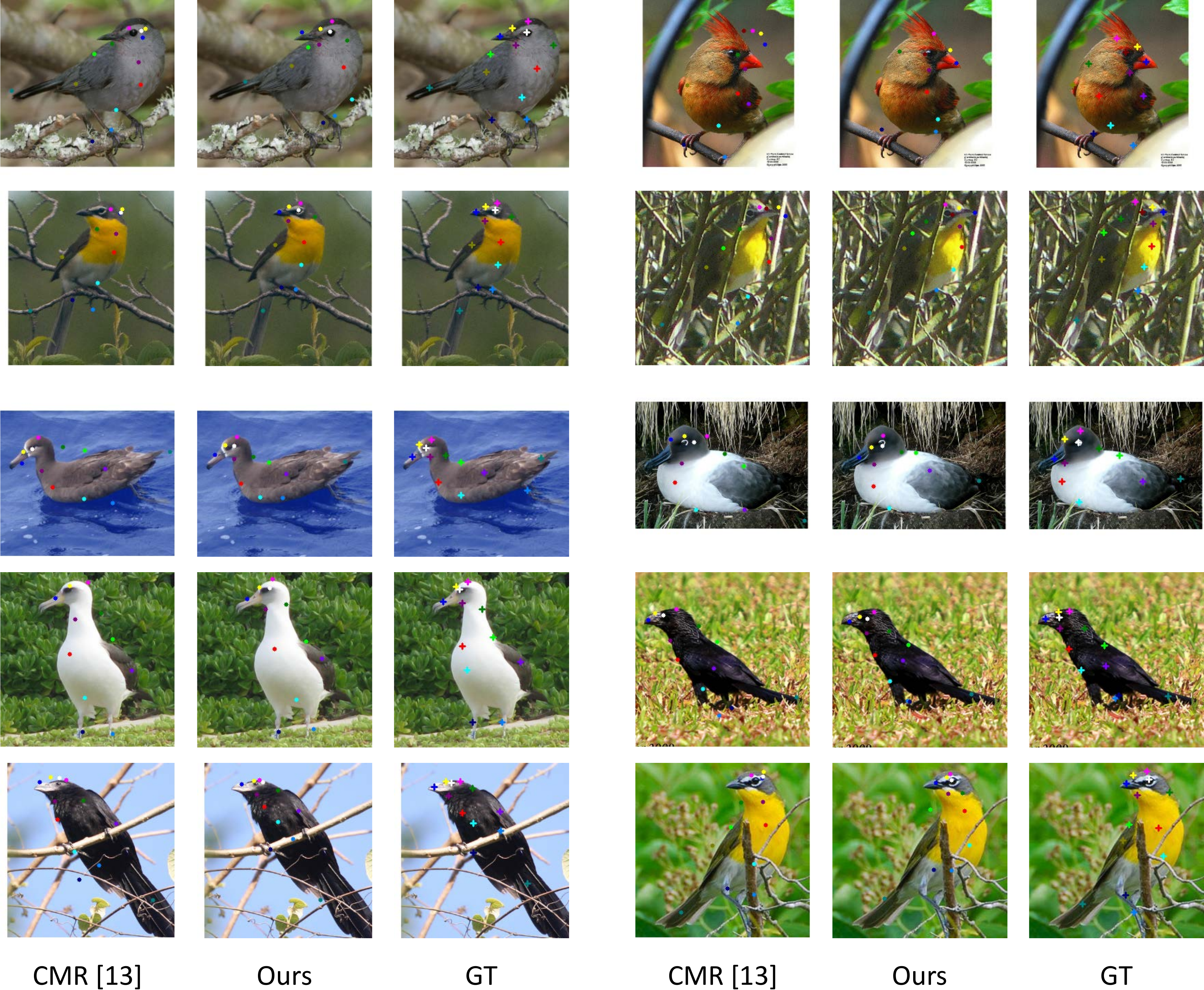}
  \caption{\footnotesize Visualization of re-projected keypoints of the single-view image reconstruction model.}\label{fig:image_kp}
  \vspace{-7pt}
\end{figure}

We visualize re-projected keypoints on test images in Fig.~\ref{fig:image_kp}, where the corresponding quantitative results are presented in Sec. 4.3, Table 1 of the main paper. The proposed ACMR model is able to predict more accurate keypoints compared to the CMR~\cite{cmrKanazawa18} method, especially when the bird performs an asymmetric pose, e.g. first row in Fig.~\ref{fig:image_kp}.

\begin{figure}[h]
  \centering
    \includegraphics[width=0.9\textwidth]{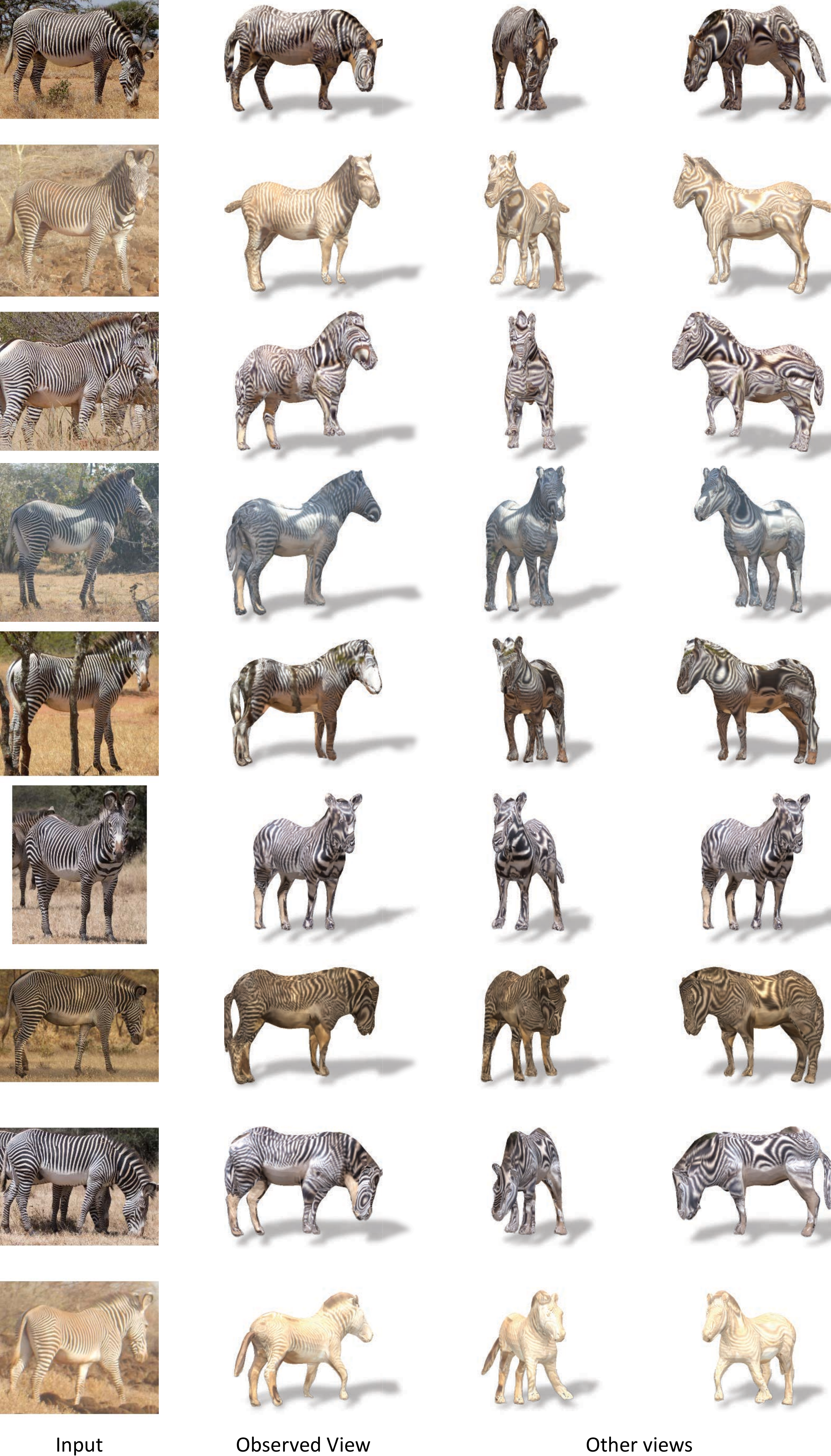}
  \caption{\footnotesize Visualization of reconstructed zebras. }\label{fig:zebra}
  \vspace{-7pt}
\end{figure}

\begin{wrapfigure}{r}{0.56\textwidth}
   \vspace{-15pt}
  \centering
    \includegraphics[width=0.53\textwidth]{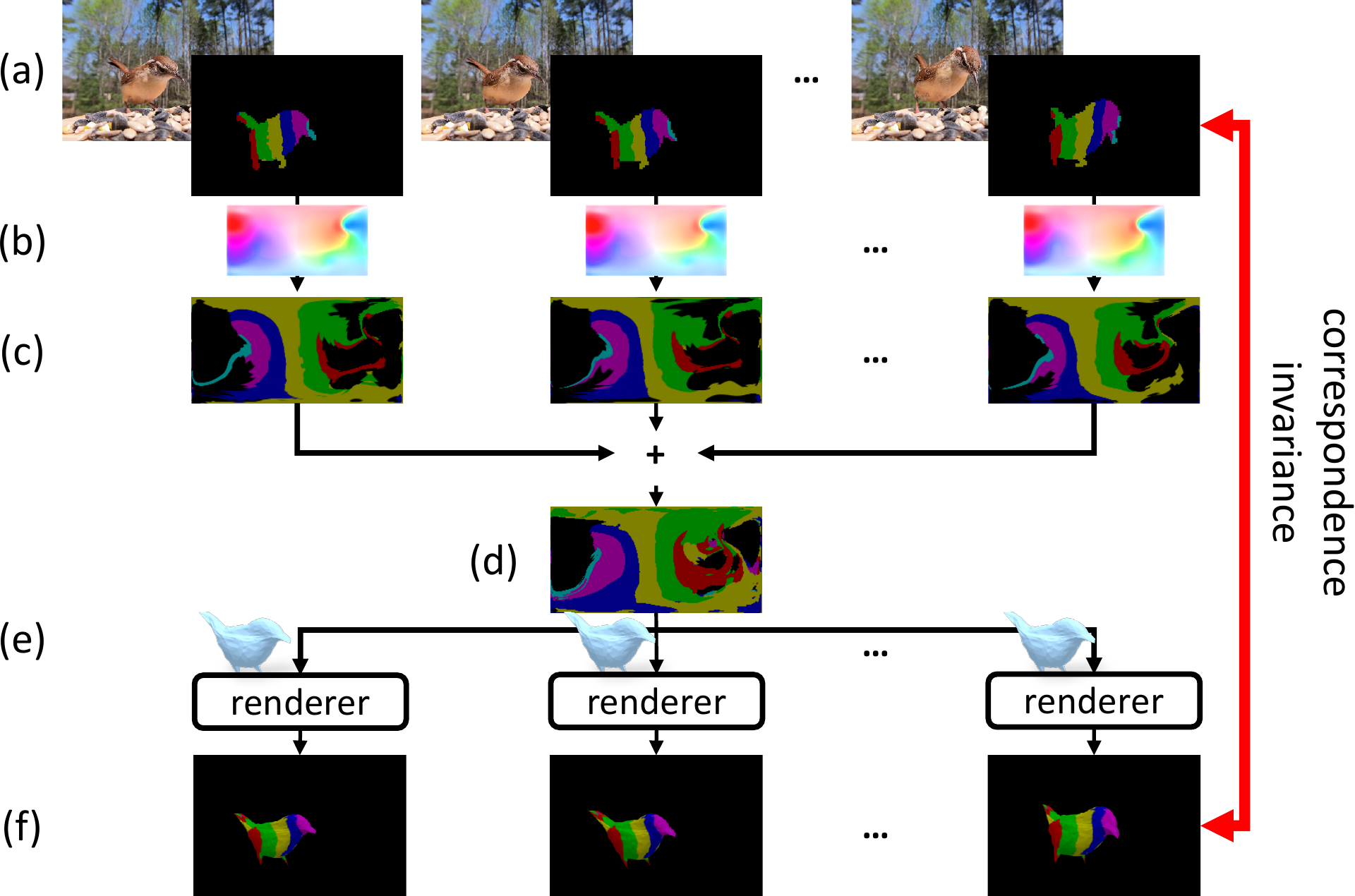}
  \caption{\footnotesize Part correspondence constraint. (a) Input frame and part propagations. (b) Predicted texture flows. (c) Part UV map. (d) Aggregated video-level part UV map. (e) Base shape and differentiable renderer. (f) Part rendering.}\label{fig:part_cons}
  \vspace{-14pt}
\end{wrapfigure}

\subsection{Single-view Image Reconstructions}
In Fig.~\ref{fig:image_quali}, we show more visualizations of reconstructions from single-view images of the test dataset of CUB birds~\cite{wah2011caltech} as well as comparisons with the baseline method~\cite{cmrKanazawa18}. By removing the symmetric assumption, our model is able to reconstruct objects in the input images more faithfully versus the baseline method~\cite{cmrKanazawa18} (Fig.~\ref{fig:image_quali}(g)).

We also demonstrate the effectiveness of the ARAP constraint, as discussed in Sec.3.1 in the paper. Without this constraint, the reconstructed meshes contain unnatural spikes, as shown in Fig.~\ref{fig:image_quali}(f).

Finally, we show reconstruction results of zebra images of the test dataset~\cite{Zuffi19Safari} in Fig.~\ref{fig:zebra}. Our ACMR model successfully captures motions such as head bending or walking for zebras.

\vspace{-2mm}
\section{Network Architecture}
\vspace{-1mm}
\label{asec:net_arch}

\subsection{Bases Visualization}
\begin{figure}[h]
  \centering
    \includegraphics[width=0.98\textwidth]{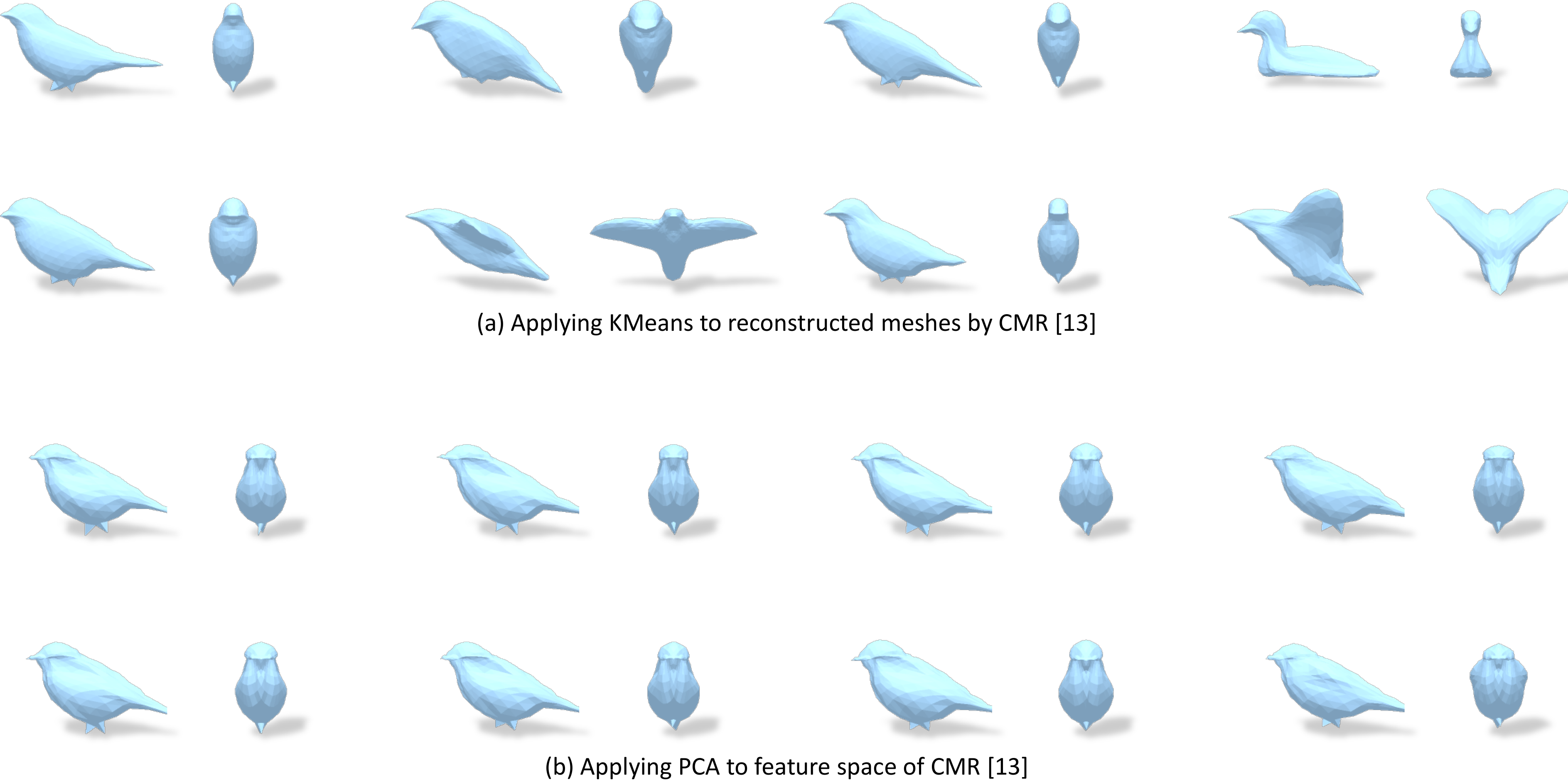}
  \caption{\footnotesize Bases visualization.}\label{fig:bases}
  \vspace{-7pt}
\end{figure}

We visualize the eight shape bases obtained by applying KMeans clustering on all reconstructed meshes by the CMR~\cite{cmrKanazawa18} method for birds in Fig.~\ref{fig:bases}(a). We also show the bases obtained by applying PCA to the bottleneck features of the image encoder. Note that the latter fails to discover rare shape modes (e.g., duck and flying bird) in the dataset as shown in Fig.~\ref{fig:bases}(b). Thus we choose to use KMeans to obtain shape bases.

\subsection{Part Correspondence Constraint}

We illustrate the part correspondence constraint in details in Fig.~\ref{fig:part_cons}. Given the propagated parts in each frame in Fig.~\ref{fig:part_cons}(a), we map them to the UV space with the predicted texture flow in Fig.~\ref{fig:part_cons}(b) and obtain part UV maps in Fig.~\ref{fig:part_cons}(c). By aggregating these part UV maps, i.e., averaging, we minimize noise in each individual part UV map and obtain a video-level part UV map in Fig.~\ref{fig:part_cons}(d). This video-level part UV map is shared by all frames in the video. Thus, for each frame, we wrap the video-level part UV map onto the base shape prediction and render it under the predicted camera pose as shown in Fig.~\ref{fig:part_cons}(f). Finally, we encourage consistency between part renderings and part propagations, as shown by the red arrow in Fig.~\ref{fig:part_cons}. Through the differentiable renderer, the loss implicitly improves the predicted camera pose.

\subsection{Single-view Reconstruction Network}
\begin{figure}[h]
  \centering
    \includegraphics[width=0.98\textwidth]{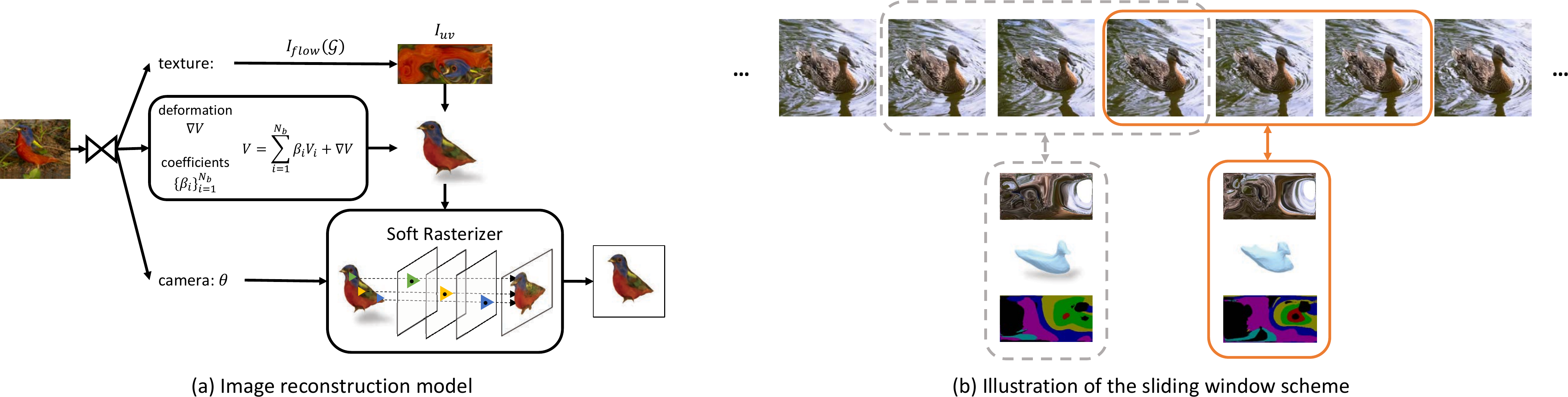}
  \caption{\footnotesize Frameworks. For the purposes of illustration only, we show the test-time tuning procedure in (b) with a sliding widow size of 3 and sliding stride of 2. In all our experiments, we use a sliding window size of 50 and stride of 10. The gray dashed box shows the previous sliding window while the orange box shows the current sliding window.}\label{fig:arch}
  \vspace{-7pt}
\end{figure}

In Fig.~\ref{fig:arch}(a), we show details of our single-view reconstruction network. Given an input image, the network jointly predicts texture, shape and camera pose. By utilizing a differentiable renderer~\cite{liu2019softras}, we are able to utilize 2D supervision, i.e. silhouettes and input images.

\subsection{Sliding Window Scheme}
We show the proposed test-time tuning process in Fig.~\ref{fig:arch}(b). Within each sliding window, we encourage the consistency of UV texture, UV parts as well as base shape of all frames.

\vspace{-2mm}
\section{Implementation Details}
\vspace{-1mm}
\label{asec:training_details}

\subsection{Objectives for Image Reconstruction Model}
\label{asec:obj_img}
We summarize the objectives used in the single-view reconstruction model (Fig.~\ref{fig:arch}(a)) discussed in Sec.3.1 in the submission as follows: (i) \textit{foreground mask loss:} a negative intersection over union objective between rendered and ground truth silhouettes~\cite{cmrKanazawa18,umr2020,kato2019vpl}; (ii) \textit{foreground RGB texture loss:} a perceptual metric~\cite{cmrKanazawa18,umr2020,zhang2018perceptual} between rendered and input RGB images; (iii) \textit{mesh smoothness: } a Laplacian constraint~\cite{cmrKanazawa18,umr2020} to encourage smooth mesh reconstruction; (iv) \textit{keypoint re-projection loss:} as discussed in Sec.3.1 in the paper; and (v) \textit{the ARAP constraint:} described in Sec.3.1 in the paper. The weight for each objective is set to 3.0, 3.0, 0.0008, 5.0 and 10.0.

\subsection{Objectives for Online Adaptation}
\label{asec:obj_video}
We summarize the objectives used in the online adaptation process (Fig.~\ref{fig:arch}(b)) in the following. Since it is feasible to predict a segmentation mask via a pretrained segmentation model, we make use of the predicted foreground mask and compute the (i), (ii), and (iii) losses (mentioned above) similarly to the image-based training.
%
We also adopt the the ARAP constraint described in Sec.3.1 in the paper, and the three invariance constraints as discussed in Sec.3.2 for online adaptation. 
%
The weight for each objective is set to 0.1, 0.5, 0.0006, 2.0 and 2.0 (texture invariance), 1.0 (part correspondence), 1.0 (base shape invariance).

\subsection{Training Details}
We implement the proposed method in PyTorch~\cite{NEURIPS2019_9015} and use the Adam optimizer~\cite{kingma2014adam} with a learning rate of 0.0001 for both the image reconstruction model training and online adaptation. The weight of each objective for the image reconstruction model as well as the online adaptation process is discussed in Sec.~\ref{asec:obj_img} and Sec.~\ref{asec:obj_video}, respectively.

\subsection{Training on Zebra Images and Videos}
We adopt a different scheme to train a single-view reconstruction model on zebras: (i) since natural zebra images labeled with keypoints are not publicly available, we adopt a synthetic dataset~\cite{Zuffi19Safari}, (ii) zebras have more complex shapes with large concavities. Therefore, it is not suitable to learn the shape by deforming from a sphere primitive. Instead, we utilize a readily available zebra mesh as a template and learn motion deformation on top of it.
We first train an image reconstruction model using the synthetic dataset provided by~\cite{Zuffi19Safari}. Similarly as~\cite{Zuffi19Safari}, we utilize the silhouettes, keypoints, texture maps as well as partially available UV texture flow as supervision.
For shape reconstruction, instead of the utilizing the SMAL parametric model~\cite{Zuffi:CVPR:2017}, we use the proposed shape module, i.e. combination of base shapes and motion deformation. Due to the limited motion of zebras, we only use one base shape, which is a readily available zebra mesh with 3889 vertices and 7774 faces.
For camera pose prediction, we use the perspective camera pose discussed in Sec.3 in the submission as well as in~\cite{cmrKanazawa18}.
Due to the limited capacity of a single UV texture map, we also model the texture map by cutting the UV texture map into four pieces and stitch them together similarly as in~\cite{Zuffi19Safari}. We note that this ``cutting and stitching'' operation does not influence the mapping and aggregation of the part UV maps discussed in Sec.3.2 in the submission.

\vspace{-2mm}
\section{Failure Cases}
\begin{figure}[h]
  \centering
    \includegraphics[width=0.98\textwidth]{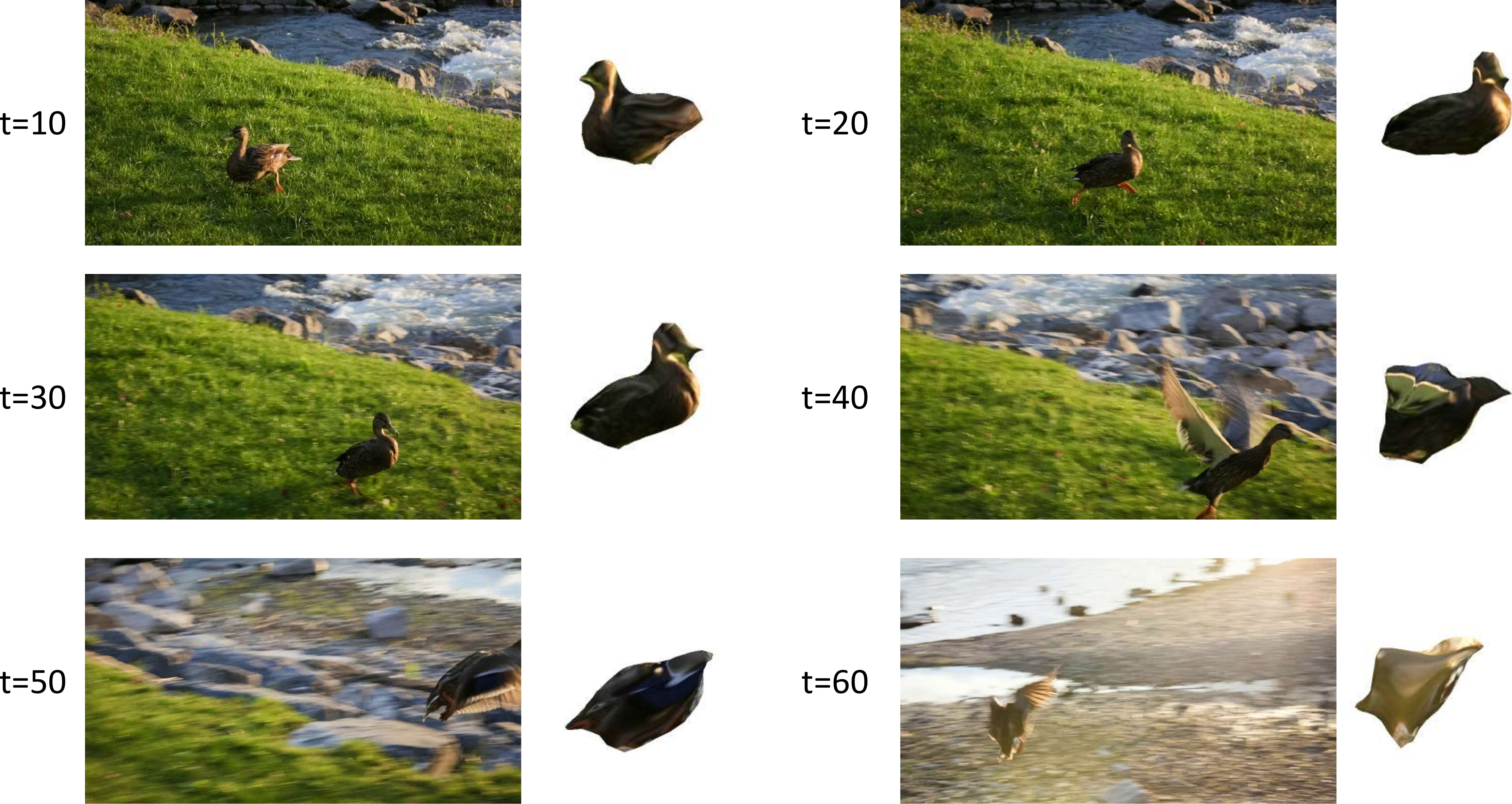}
  \caption{\footnotesize Failure cases. Our model fails when there is occlusion (e.g., $t=50$, $t$ stands for frame number) or large lighting changes (e.g., $t=60$).}\label{fig:failure}
  \vspace{-7pt}
\end{figure}

\vspace{-1mm}
\label{asec:failure_cases}
Our work is the first to explore the challenging task of reconstructing 3D meshes of deformable object instances from videos in the wild. Impressive as the performance is, this challenging task is far from being fully solved. We discuss failure cases and limitations of the proposed method in the following.
To begin with, we focus on genus-0 objects such as birds and zebras in this work. Thus our model suffers when it is generalized to objects with large concave holes such as chairs, humans etc.
Second, our work struggles to reconstruct meshes from videos with large motion and lighting changes as well as occlusion, (see Fig.~\ref{fig:failure}). This is mainly due to the failure in correctly propagating parts by the self-supervised UVC model~\cite{uvc_2019}, which is out of  scope of this work. 
We leave all these failure cases and limitations to future works.

{\small
\bibliographystyle{ieee}
\bibliography{vmr}
}
\end{document}